%% file: example_paper.tex
\PassOptionsToPackage{dvipsnames,table}{xcolor}
\documentclass{article}

\usepackage{microtype}
\usepackage{graphicx}
\usepackage{subfig}
\usepackage{booktabs} 

\usepackage{hyperref}

\usepackage[accepted]{icml2024}

\usepackage{amsmath}
\usepackage{amssymb}
\usepackage{mathtools}
\usepackage{amsthm}

\usepackage[capitalize,noabbrev]{cleveref}

\theoremstyle{plain}

\theoremstyle{definition}

\theoremstyle{remark}

\usepackage[textsize=tiny]{todonotes}

\usepackage{times}
\usepackage{latexsym}
\usepackage{natbib}

\usepackage[T1]{fontenc}

\usepackage[utf8]{inputenc}

\usepackage{microtype}
\usepackage{url}
\usepackage{amsmath,amsfonts,amssymb}

\usepackage{multirow}
\usepackage{mathrsfs}
\usepackage{algorithm}
\usepackage{algorithmic}
\usepackage{graphicx}
\usepackage{bbm}
\usepackage{wrapfig,lipsum,booktabs}
\usepackage{pifont}
\usepackage{placeins}
\usepackage[normalem]{ulem}

\makeatletter
\patchcmd{\BR@backref}{\newblock}{\newblock(page~}{}{}
\patchcmd{\BR@backref}{\par}{)\par}{}{}
\makeatother

\useunder{\uline}{\ul}{}
\newcommand{\cmark}{\ding{51}}
\newcommand{\xmark}{\ding{55}}

\newcommand{{\ourmethod}}{APT}
\newcommand{{\ourarch}}{APT adapter}
\newcommand{\ourarchabbr}{APT adapter}
\newcommand{\tableindent}{~~~~}
\newcommand{\lmabbr}{LM}

\icmltitlerunning{APT: Adaptive Pruning and Tuning Pretrained Language Models for Efficient Training and Inference}

\begin{document}

\twocolumn[
\icmltitle{APT: Adaptive Pruning and Tuning Pretrained Language Models for \\Efficient Training and Inference}



\icmlsetsymbol{equal}{*}

\begin{icmlauthorlist}
\icmlauthor{Bowen Zhao}{xxx}
\icmlauthor{Hannaneh Hajishirzi}{xxx,yyy}
\icmlauthor{Qingqing Cao$^*$}{zzz}
\end{icmlauthorlist}

\icmlaffiliation{xxx}{University of Washington}
\icmlaffiliation{yyy}{Allen Institute for Artificial Intelligence}
\icmlaffiliation{zzz}{$^*$Apple, work done at the University of Washington}

\icmlcorrespondingauthor{Bowen Zhao}{bowen98@uw.edu}
\icmlcorrespondingauthor{Qingqing Cao}{qicao@apple.com}

\icmlkeywords{Machine Learning, ICML}

\vskip 0.3in
]



\printAffiliationsAndNotice{}  

\begin{abstract}
Fine-tuning and inference with large Language Models~({\lmabbr}) are generally known to be expensive. 
Parameter-efficient fine-tuning over pretrained LMs reduces training memory by updating a small number of {\lmabbr} parameters but does not improve inference efficiency. Structured pruning improves {\lmabbr} inference efficiency by removing consistent parameter blocks, yet often increases training memory and time. To improve both training and inference efficiency, we introduce {\ourmethod} that adaptively {\it prunes} and {\it tunes} parameters for the {\lmabbr}s. 
At the early stage of fine-tuning, {\ourmethod} dynamically adds {\it salient} tuning parameters for fast and accurate convergence while discarding unimportant parameters for efficiency.
Compared to baselines, our experiments show that {\ourmethod} maintains up to 98\% task performance when pruning 60\% of the parameters in RoBERTa and T5 models. APT also preserves 86.4\% of LLaMA models' performance with 70\% parameters remaining. Furthermore, {\ourmethod} speeds up LMs' fine-tuning by up to 8$\times$ and reduces large {\lmabbr}s' memory training footprint by up to 70\%. Our code and models are publicly available at \url{https://github.com/ROIM1998/APT}.
\end{abstract}

\input{sections/introduction}
\input{sections/background}
\input{sections/method}
\input{sections/experiment}
\input{sections/conclusion}

\section*{Acknowledgements}
This research was supported partly by NSF IIS-2044660, an Allen Investigator Distinguished award. We thank the members of the UW NLP group for their comments and feedback on this paper.

\input{sections/impact_statement}


\bibliography{official_custom}
\bibliographystyle{icml2024}

\newpage
\appendix
\onecolumn

\input{sections/appendix}


\end{document}

%% file: sections/introduction.tex
\section{Introduction}
Fine-tuning language models (LMs)~\citep{devlin-etal-2019-bert,liu2019roberta,raffel2020exploring} is an essential paradigm to adapt them to downstream tasks~\citep{mishra-etal-2022-cross,wang-etal-2022-super}. Increasing the parameter scale of LMs improves model performance~\citep{kaplan2020scaling}, but incurs significant training and inference costs.  
For instance, a 13B LLaMA model~\citep{touvron2023llama} costs about 100GB memory for fine-tuning and 30GB for inference with float16 datatype. 
It is important to improve the training and inference efficiency of {\lmabbr} for practical applications.

\begin{figure}[t!]
    \centering
    \includegraphics[width=0.9\linewidth]{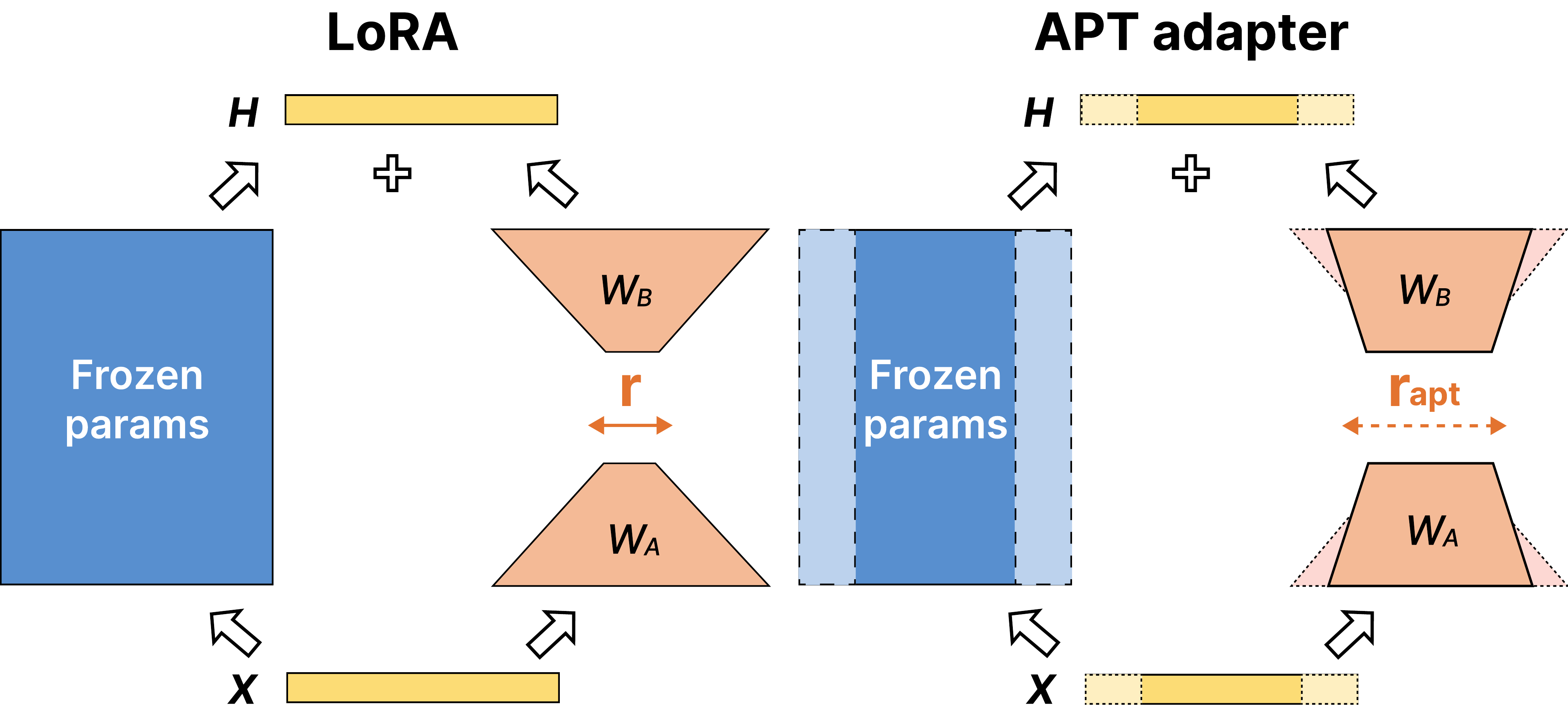}
\caption{{\ourmethod} provides both training and inference efficiency benefits by pruning and tuning pretrained LM parameters adaptively via the \textbf{{\ourmethod} adapter}. We dynamically adjust (add/reduce) {\ourmethod} adapter input/output dimensions and the rank ($r_{\text{apt}}$). Reducing adapter dimensions prunes frozen parameters, making training and inference faster and more memory-efficient. Adding adapter ranks helps recover the pruned {\lmabbr}'s task performance. In contrast, existing adapters like LoRA allow efficient training but do not provide inference efficiency since the model size is not reduced.}
    \label{fig:teaser}
    \vspace{-15pt}
\end{figure}


\input{tables/preliminary_summary}

Parameter-efficient fine-tuning methods (PEFT, summarized in \cref{tab:preliminary-summary})~\citep{houlsby2019parameter,li2021prefix} reduce the memory consumption of LM fine-tuning via updating a small number of parameters. However, PEFT models do not improve inference efficiency because the {\lmabbr} size remains the same or even increases after fine-tuning. For instance, LoRA~\citep{hu_lora_2021} tunes low-rank decomposed linear layers parallel to frozen parameters to reduce training memory but takes longer to converge~\citep{ding2023parameter}.
On the other hand, structured pruning~\cite{kwon_fast_2022,xia_structured_2022,ma2023llm} improves inference efficiency by removing blocks of parameters such as attention heads and feed-forward neurons in Transformer {\lmabbr}s, showing more inference speedup than sparse unstructured pruning methods~\citep{Han2015DeepCC,han2015learning,sanh_movement_2020}. However, training pruned {\lmabbr}s takes extra time to converge and incurs high memory, substantially diminishing LMs' accessibility in usage scenarios with limited computational resources.

Integrating structured pruning and PEFT could increase both training and inference efficiency. 
However, existing research~\citep{zhao2023cpet} indicates that combining PEFT and structured pruning, such as applying structured pruning over LoRA-tuned models, causes noticeable performance loss and extra training costs. It remains challenging to prune {\lmabbr}s accurately using limited training resources.

In this paper, we develop an efficient fine-tuning approach named {\ourmethod} that \textbf{A}daptively selects model parameters for \textbf{P}runing and fine-\textbf{T}uning. {\ourmethod} combines the benefits of PEFT and structured pruning to make fine-tuning and inference more efficient. Our intuition is that pre-trained {\lmabbr} parameters contain general knowledge, but their importance to downstream tasks varies. Therefore, we can remove the parameters irrelevant to the fine-tuning task in the early training stage. Early-removing these parameters improves training and inference efficiency while not substantially hurting model accuracy~\citep{frankle2021pruning,Shen_2022_CVPR,zhang2023emergent}. Meanwhile, continuously adding more parameters for fine-tuning can improve LM performance because task-specific skills live in a subset of {\lmabbr} parameters~\citep{wang-etal-2022-finding-skill,panigrahi2023task}. 

More specifically, {\ourmethod} learns the pruning masks via an outlier-aware salience scoring function to remove irrelevant {\lmabbr} parameter blocks and adds more tuning parameters during fine-tuning according to tuning layer importance. To make training more efficient, the salience scoring function is lightweight and causes little runtime and memory overhead. Combined with our self-distillation technique that shares teacher and student parameters, {\ourmethod} can accurately prune an {\lmabbr} with less training time and lower memory usage.
Experimental results show that {\ourmethod} prunes RoBERTa and T5 base models 8$\times$ faster than the LoRA plus pruning baseline while reaching 98.0\% performance with 2.4 $\times$ speedup and 78.1\% memory consumption during inference. When pruning large {\lmabbr}s like LLaMA, {\ourmethod} costs only 30\% memory compared to the state-of-the-art pruning method and still maintains 86.4\% performance with 70\% parameters. Our ablation study in \cref{sec:ablation} indicates the effectiveness of adaptive pruning and tuning. It also demonstrates that efficient distillation with {\ourarchabbr} substantially recovers small {\lmabbr}s' performance while outlier-aware salience scoring prunes large {\lmabbr}s more accurately. Our analysis in \cref{sec:analysis} demonstrates that controlled adaptive tuning with early pruning during fine-tuning improves {\lmabbr} end-task accuracy better with less training time and memory costs.

%% file: tables/preliminary_summary.tex
\begin{table*}[t!]
\centering
\small
\begin{tabular}{@{}ll|cc|cc|cc@{}}
\toprule
\multicolumn{2}{l|}{\multirow{2}{*}{Method}} & \multirow{2}{*}{$\mathcal{A}_{\text{P}}$} & \multirow{2}{*}{$\mathcal{A}_{\text{T}}$} & \multicolumn{2}{c|}{Training}                                             & \multicolumn{2}{c}{Inference}                                                        \\
\multicolumn{2}{l|}{}                        &                     &                     & T                                          & M                         & T                                                  & M                          \\ \midrule
\multirow{3}{*}{PEFT}       & Adapter~\citep{Pfeiffer2020AdapterFusionNT}        & \xmark              & \xmark              & \color{red}$\Uparrow_{\text{High}}$ & \color{ForestGreen}$\Downarrow_{\text{Low}}$ & \color{red}$\Uparrow_{\text{Low}}$                              & \color{red}$\Uparrow_{\text{Low}}$      \\
                            & LoRA~\citep{hu_lora_2021}           & \xmark              & \xmark              & \color{red}$\Uparrow_{\text{High}}$ & \color{ForestGreen}$\Downarrow_{\text{Low}}$ & \textbf{=}                                         & \textbf{=}                 \\
                            & AdaLoRA~\citep{zhang2023adaptive}        & \xmark              & \cmark              & \color{red}$\Uparrow_{\text{High}}$ & \color{ForestGreen}$\Downarrow_{\text{Low}}$ & \textbf{=}                                         & \textbf{=}                 \\ \midrule
\multirow{4}{*}{Pruning}    & MvP~\citep{sanh_movement_2020}            & \xmark              & \xmark              & \color{red}$\Uparrow_{\text{High}}$ & \color{red}$\Uparrow_{\text{Low}}$     & \color{ForestGreen}$\Downarrow_{\text{Low}}$                         & \color{ForestGreen}$\Downarrow_{\text{Low}}$ \\
                            & BMP~\citep{lagunas_block_2021}            & \xmark              & \xmark              & \color{red}$\Uparrow_{\text{High}}$ & \color{red}$\Uparrow_{\text{Low}}$     & \color{ForestGreen}$\Downarrow_{\text{High}}$ & \color{ForestGreen}$\Downarrow_{\text{Low}}$  \\
                            & CoFi~\citep{xia_structured_2022}           & \xmark              & \xmark              & \color{red}$\Uparrow_{\text{High}}$ & \color{red}$\Uparrow_{\text{Low}}$     & \color{ForestGreen}$\Downarrow_{\text{High}}$ & \color{ForestGreen}$\Downarrow_{\text{Low}}$  \\
                            & MT~\citep{kwon_fast_2022}             & \xmark              & \xmark              & \textbf{=}                                 & \textbf{=}                & \color{ForestGreen}$\Downarrow_{\text{High}}$ & \color{ForestGreen}$\Downarrow_{\text{Low}}$  \\ \midrule
\multirow{3}{*}{Combined}   & SPA~\citep{hedegaard_structured_2022}            & \xmark              & \xmark              & \color{red}$\Uparrow_{\text{High}}$ & \color{red}$\Uparrow_{\text{Low}}$     & \color{ForestGreen}$\Downarrow_{\text{High}}$ & \color{ForestGreen}$\Downarrow_{\text{Low}}$  \\
                            & LRP~\citep{zhang2023pruning}            & \xmark              & \xmark              & \color{red}$\Uparrow_{\text{High}}$ & \color{ForestGreen}$\Downarrow_{\text{Low}}$ & \color{ForestGreen}$\Downarrow_{\text{High}}$ & \color{ForestGreen}$\Downarrow_{\text{Low}}$  \\
                            & \textbf{\ourmethod} (ours)   & \cmark              & \cmark              & \color{red}$\Uparrow_{\text{Low}}$                      & \color{ForestGreen}$\Downarrow_{\text{Low}}$ & \color{ForestGreen}$\Downarrow_{\text{High}}$ & \color{ForestGreen}$\Downarrow_{\text{Low}}$  \\ \bottomrule
\end{tabular}
\caption{Efficiency comparison of existing methods and APT. $\mathcal{A}_{\text{P}}$ stands for adaptive pruning and $\mathcal{A}_{\text{T}}$ for adaptive tuning, where the total and tuning parameter sizes are dynamically adjusted. We measure efficiency using training converge time, inference time (T), and peak memory (M). Symbols {\color{red}$\Uparrow$} and {\color{ForestGreen}$\Downarrow$} indicate increased and decreased costs, respectively, while \textbf{=} signifies no change in cost. The terms ``low'' and ``high'' qualify the extent of cost variations.} 
\label{tab:preliminary-summary}
\vspace{-15pt}
\end{table*}

%% file: sections/background.tex
\section{Related Works} 

\subsection{Parameter-efficient Fine-tuning (PEFT)}
PEFT methods aim to tune {\lmabbr}s with limited resources by updating a small number of parameters~\citep{lialin2023scaling}, mainly falling into three categories: selective, additive, and dynamic. Selective methods focus on tuning a subset of parameters in {\lmabbr}s with pre-defined rules~\citep{ben-zaken-etal-2022-bitfit} or importance metrics~\citep{sung2021training,guo-etal-2021-parameter}. Additive methods tune injected layer modules~\citep{houlsby2019parameter,Pfeiffer2020AdapterFusionNT} or embeddings~\citep{lester-etal-2021-power,li2021prefix}. For example, LoRA~\citep{hu_lora_2021} tunes low-rank decomposed layers to avoid inference cost overhead. However, LoRA keeps the tuning layer shapes static without dynamic adjustments. Dynamic methods~\citep{he-etal-2022-sparseadapter} adjust tuning parameters during training. For instance, AdaLoRA~\citep{zhang2023adaptive} gradually reduces tuning parameters but does not benefit inference efficiency. Compared to these methods, {\ourmethod} adaptively adjusts the pruning and tuning parameters simultaneously, improving training and inference efficiency.

\subsection{Model Compression}
Model compression methods like quantization and pruning boost inference efficiency. Quantization aims to reduce {\lmabbr}s' memory consumption via converting parameters to low-bit data types~\citep{Frantar2022GPTQAP,Dettmers2022LLMint88M,lin2023awq}. However, despite reducing {\lmabbr}'s memory consumption, the speedup benefits of quantization require specific framework support, which limits their adaptability.
Pruning~\citep{LeCun1989OptimalBD,Han2015DeepCC,frankle2018lottery,xu2021rethinking} aims to discard unimportant parameters in {\lmabbr}s for inference efficiency. Unstructured pruning~\citep{sanh_movement_2020} prunes sparse parameters in {\lmabbr}s, which requires dedicated hardware support for efficiency improvements. Meanwhile, structured pruning~\citep{lagunas_block_2021,xia_structured_2022} prunes consistent blocks in transformer layers (MHA heads, FFN neurons, and model dimensions) for ubiquitous inference efficiency gains. Such pruning often uses knowledge distillation~\citep{Hinton2015DistillingTK}, which causes more training costs. Post-training pruning~\citep{kwon_fast_2022,Frantar2023SparseGPTML} aims to prune fine-tuned models with limited extra costs but requires initialization from fully fine-tuned models. Moreover, task-agnostic pruning~\citep{sun2023simple,ma2023llm} cannot achieve on-par performance with task-specific pruning.

\subsection{Combining Compression and PEFT}
Combining model compression and PEFT might achieve both training and inference efficiency improvements: QLoRA~\citep{dettmers2023qlora} and QA-LoRA~\citep{xu2023qa} bring quantization and LoRA together for large {\lmabbr} tuning. SPA~\citep{hedegaard_structured_2022} combines structured pruning and Compacter~\citep{karimi2021compacter}, yet suffers substantial performance loss. CPET~\citep{zhao2023cpet} leverages different task-agnostic model compression methods together with LoRA and knowledge distillation, but the performance loss becomes notable specifically when structured pruning is applied. PST~\citep{ijcai2022p586} and LRP~\citep{zhang2023pruning} also explored the combination of LoRA and pruning, yet their performance degradations are also substantial because their tuning parameters are static.
In contrast, {\ourmethod} identifies tuning and pruning parameters based on their salience in fine-tuning, which can improve training and inference efficiency under a new paradigm with minimal performance loss. 

%% file: sections/method.tex
\section{Problem Formulation} \label{sec:formulation}

Our goal is to improve the training and inference efficiency of pretrained {\lmabbr} while maintaining task performance. Intuitively, tuning fewer parameters leads to smaller training memory footprints and shorter time per training step; models with fewer parameters also run faster with less memory footprint during inference but come with task performance degradation. We aim to find the optimal parameters for training and inference without sacrificing task performance. 

We formally define the problem objective as minimizing the task loss $\mathcal{L}$ under the constraint that the total {\lmabbr} parameter size $\Theta$ reaches a target sparsity (defined as the ratio of the number of parameters pruned to the original {\lmabbr}) $\gamma_T$ after $T$ training steps. For each training step $t$, the sparsity of the {\lmabbr} remains above $\gamma_t$ while the number of tuning parameters is below $\Delta_t$. We control the pruning masks $\mathcal{M}_t$ and tuning ranks $\mathcal{R}_t$ to satisfy these constraints. We describe the optimization process as:
\begin{equation}
\label{eq:problem}
    \begin{aligned}
        &\operatorname*{argmin}_{\Theta_T, \mathcal{M}_T} & & \frac{1}{|\mathcal{D}|} \sum_{x, y \in \mathcal{D}}\mathcal{L}(x, y|\Theta_T, \mathcal{M}_T) \\
        & \text{s.t. }
        & & 1 - \frac{\mathcal{C}(\Theta_t, \mathcal{M}_t)}{\mathcal{C}(\Theta_0, \mathcal{M}_0)} \geq \gamma_t, \\
        &&& \delta(\Theta_t, \mathcal{M}_t, \mathcal{R}_t) \leq \Delta_t, \\
        &&& \forall t \in \{0, 1, \ldots, T\}.
    \end{aligned}
\end{equation}
where $x, y$ are inputs and labels sampled from the task dataset $\mathcal{D}$, while $\mathcal{C}$ and $\delta$ denotes total and tuning parameter numbers of the {\lmabbr}, respectively.


Based on \cref{eq:problem}, a higher target sparsity $\gamma_T$ improves inference efficiency with fewer FLOPs and memory usage but sacrifices performance. Increasing $\gamma_t$ when $t \ll T$ also improves training efficiency. Besides, tuning more parameters with larger $\Delta$ costs more training memory but makes the model converge faster with better task performance. Our formulation supports task performance improvements together with training and inference efficiency by dynamically adjusting the LM parameters during fine-tuning. 

\section{Adaptive Pruning and Tuning}

\begin{figure*}[t!]
    \centering
    \includegraphics[width=0.95\textwidth]{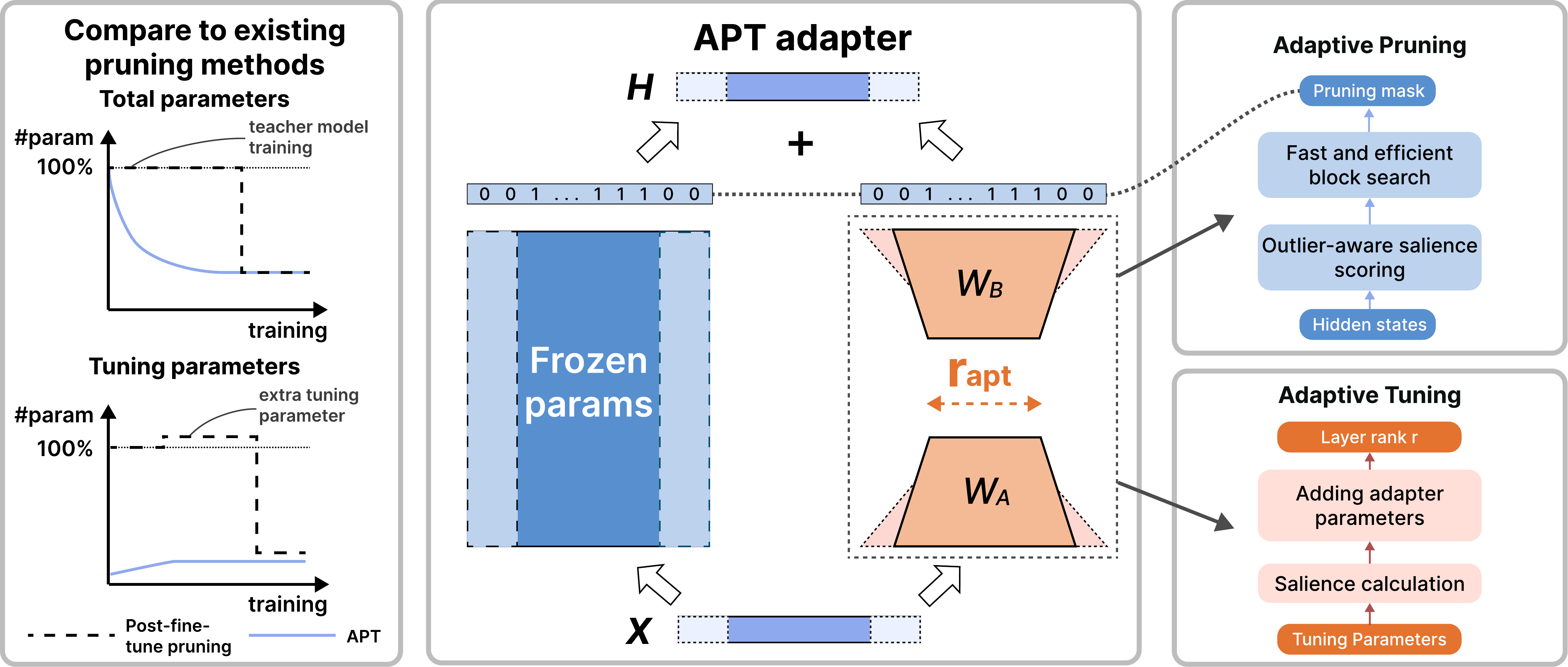}
    \caption{{\ourmethod} adaptively identifies pruning and tuning parameters via {\ourarchabbr}s during fine-tuning with little cost. {\ourmethod} gradually prunes {\lmabbr} parameters with binary pruning masks learned from our lightweight outlier-aware salience scoring function for training and inference efficiency. {\ourmethod} also adds tuning parameters in salient layers in {\lmabbr} fine-tuning through increasing dynamic ranks in {\ourarchabbr}s for performance recovery.}
    \label{fig:elastic-adapter}
\vspace{-15pt}
\end{figure*}

We design \textbf{A}daptive \textbf{P}runing and \textbf{T}uning (\textbf{APT}) over {\lmabbr} parameters to allow efficient training and inference while maintaining task performance. 

Summarized in the left of \cref{fig:elastic-adapter}, existing pruning methods often neglect training costs where the number of tuning parameters is more than a parameter-efficient threshold with $\Delta_t \ge \mathcal{C}(\Theta_t, \mathcal{M}_t)$, resulting in long training time and high memory consumption. Instead, to improve training efficiency, we prune {\lmabbr} parameters (increase $\gamma_t$) during early training when $t \ll T$ while keeping $\Delta_t \ll \mathcal{C}(\Theta_t, \mathcal{M}_t)$ to reduce training costs. In addition, we add tuning parameters (increase $\Delta_t$) in early training to effectively mitigate the degradation of {\lmabbr}'s performance due to pruning.



\noindent \textbf{Overview.} \cref{fig:elastic-adapter} shows the overview of our method that incorporates our new APT adapter for pruning and tuning. Our intuition is that pruning {\lmabbr}s during early fine-tuning will not hurt their task performance while reducing training and inference costs. Meanwhile, unlike existing adapters like LoRA~\cite{hu_lora_2021} that use fixed tuning parameters, APT adapters dynamically add tuning parameters to accelerate {\lmabbr} convergence with superior task performance. We first introduce the architecture of APT adapters in \cref{sec:apt}. We then describe how we prune {\lmabbr} parameters at early fine-tuning with low cost in \cref{sec:apt-prune} and adaptively tune LMs to recover task performance efficiently in \cref{sec:apt-tune}. Additionally, we explain our self-knowledge distillation technique that improves pruned {\lmabbr}'s task performance with limited training expense in \cref{sec:momentum}. 

\subsection{{\ourarchabbr}} \label{sec:apt}
We build the {\ourarchabbr} architecture over LoRA, but the key difference is that {\ourarchabbr} supports dynamic {\lmabbr} pruning and tuning. Assuming an {\ourarchabbr} projects the input $X \in \mathbb{R}^{d_i}$ to the output $H_{\text{apt}}(X) \in \mathbb{R}^{d_o}$, we design binary pruning masks ($m_i \in \mathbb{R}^{d_i}$ for input and $m_o \in \mathbb{R}^{d_o}$ for output) and dynamic ranks $r_{\text{apt}}$ in {\ourarchabbr} to control the total and tuning {\lmabbr} parameters during fine-tuning, respectively. Specifically, with tuning parameters $W_A \in \mathbb{R}^{r_{\text{apt}} \times d_i}$ and $W_B \in \mathbb{R}^{d_o \times r_{\text{apt}}}$, {\ourarchabbr} $H_{\text{apt}}$ is denoted as:
\begin{equation} \label{eq:elastic-lora}
        H_{\text{apt}}(X) = m_o \circ (W + s \cdot W_B W_A )X \circ m_i
\end{equation}
where $s$ is the constant scaling factor following LoRA's implementation, and $\circ$ denotes the Hadamard product between the masks and their corresponding matrices. The parameter block is pruned when the multiplying mask is set to 0 and retained when set to 1. In the meantime, during fine-tuning, we dynamically increase $r_{\text{apt}}$ for the weight matrices $W_B$ and $W_A$. Compared to LoRA, APT adapters can be more efficient due to more adaptive pruning and tuning over LM parameters.

In transformer-based {\lmabbr} fine-tuning, we add {\ourarchabbr}s in queries and values of multi-head attention (MHA) layers. We also add {\ourarchabbr} in feed-forward network (FFN) layers when fine-tuning smaller models like RoBERTa and T5 for fast training convergence. In these cases, $m_i$ prunes transformers' hidden dimension and $m_o$ prunes attention heads in MHA and internal neurons in FFN layers. By learning the pruning masks and adjusting the ranks dynamically in the APT adapter, we can achieve the goal defined in \cref{sec:formulation} where the tuning parameter number $\delta(\Theta_t, \mathcal{M}_t, \mathcal{R}_t)$ increases to maintain task performance and the {\lmabbr} parameter size $\mathcal{C}(\Theta_t, \mathcal{M}_t)$ decreases to support more efficient training and inference. Next, we describe the adaptive pruning and tuning procedures in detail.

\subsection{Low-cost Adaptive {\lmabbr} Pruning ($\mathcal{A}_{\text{P}}$)} \label{sec:apt-prune}

To benefit the efficiency of LM training and inference, {\ourmethod} adaptively prunes LM parameters since the start of fine-tuning. The problem is finding the parameters to be pruned and discarding them without hurting training stability. Given a task, we compute the outlier-aware salience score of parameter blocks at each early-training step when $t \ll T$. Afterward, we use a fast search algorithm to determine the parameters to be pruned, and then we update their binary pruning masks accordingly. The upper-right of \cref{fig:elastic-adapter} shows this adaptive pruning procedure.


\noindent \textbf{Outlier-aware salience scoring of {\lmabbr} parameters.}
When determining the influence of pruning parameters on the {\lmabbr} performance for fine-tuning tasks, the key idea is to compute the outlier-aware salience scores of {\lmabbr} activations to consider both tuning and frozen parameters. 
In detail, salience is defined as the magnitude of parameters' weight-gradient production from previous works~\citep{sanh_movement_2020}, where
\begin{equation} \label{eq:salience}
    S(W_{i, j}) = |{W}_{i,j} \cdot \frac{\partial \mathcal{L}}{\partial {W}_{i,j}}|
\end{equation}

However, since the frozen weights' gradients are unreachable in PEFT settings, we compute the salience as the magnitude of the product of activations and their gradients.
Additionally, we compress the activation and gradients by summing along batches before production to further reduce the training memory consumption. 
On the other hand, block outlier parameters play a crucial role in task-specific capabilities, as previous quantization methods suggest~\citep{Dettmers2022LLMint88M,lin2023awq}. Such effects brought by outlier parameters will be averaged if salience is only measured on the block level. To keep more outlier parameters in the pruned {\lmabbr}s, we combine the salience score above and the kurtosis\footnote{Representing the density of the outlier in a distribution, the more the outliers are, the bigger the kurtosis will be.} of the activation together. Therefore, given the supervised finetuning dataset $\mathcal{D}_{t}$, the outlier-aware salience score $\hat{S}$ is defined as:
\begin{align}
    \begin{split}        
        \widetilde{S}_t(W_{:,j}) &= \sum_{(x, y) \in \mathcal{D}_t} \sum_{i} |\frac{\partial \mathcal{L}(x, y | \Theta_t, \mathcal{M}_t)}{\partial H_{j, i}}| \cdot \\
            &\quad \sum_{(x, y) \in \mathcal{D}_t} \sum_{i} | H_{j, i}|
    \end{split} \label{eq:salience-approx} \\
    \hat{S}((W_{:,j}) &= \widetilde{S}(W_{:,j}) + (\textrm{Kurt}(O_{j,:}))^{\frac12} \label{eq:outlier-awared-salience}
\end{align}
where $H$ is the activations in the {\lmabbr}, $\text{Kurt}(\cdot)$ stands for kurtosis, and $O_{:, j} = W_{:, j} \circ X_{j, :}^\intercal$ represents the activation. We leave details of the salience scoring in \cref{appendix:salience-unify}.

\noindent \textbf{Efficient search of {\lmabbr} block parameters.}
Given the salience calculated in \cref{eq:outlier-awared-salience}, the next step is to learn the binary pruning masks to increase the {\lmabbr} sparsity above $\gamma_t$. Intuitively, we shall prune the blocks with less salience score, which formulates a latency-saliency knapsack~\citep{NEURIPS2022_5434be94} task.
For an {\lmabbr} with $n_L$ transformer layers, where layer $i$ has $n_h^i$ MHA heads and $n_f^i$ FFN neurons, and all transformer layers' hidden dimension sizes are $d_m$, the approximated\footnote{We ignore the model's layer norm and bias terms since their sizes are small, and we do not count tuning parameters since they can be fully merged after training.} number {\lmabbr} parameter is:
\begin{equation} \label{eq:parameter-constraint}
    \mathcal{C}(\Theta_t; \mathcal{M}_t) \approx d_m \sum_{i=1}^{n_L} (4 n_h^i \cdot d_h + 2 n_f^i)
\end{equation}
where $d_h$ is the dimension per MHA head. To keep the constraint in \cref{eq:problem}, we prune MHA heads, FFN neurons, and the model hidden dimension simultaneously by reducing $n^i_h, n^i_f$, and $d_m$.
Hence, we first sort the blocks by their salience divided by the parameter number. As the parameter size monotonically increases with block quantity, we use binary search to identify the top salient blocks to be retained given the sparsity constraint $\gamma_t$. 
We leave the implementation details in \cref{appendix:binary-search} for simplicity.

\subsection{Adaptive and Efficient LM Tuning ($\mathcal{A}_{\text{T}}$)} \label{sec:apt-tune}
As using PEFT methods to fine-tune pruned {\lmabbr}s causes notable performance decrease (illustrated in \cref{tab:small-model-results} and \cref{tab:ablate-paramalloc}), we aim to dynamically add tuning parameters in {\lmabbr} fine-tuning to improve the model's end-task performance. However, since more tuning parameters will consume extra training time and memory, we want to add parameters in a controlled way, where new parameters are only added to task-sensitive {\ourmethod} adapters. As a result, we can recover pruned {\lmabbr}s' performance with reasonable training costs.
In detail, we first calculate the salience of each {\ourmethod} adapter to determine their importance. Next, we select the top-half {\ourmethod} adapters after sorting them with salience and add their parameters by increasing their $r_{\text{apt}}$.

\noindent \textbf{Salience scoring of {\ourarchabbr}.}
Since gradients of tuning parameters information are available when determining the layer salience, we can first calculate each tuning parameter's salience with \cref{eq:salience}. Then, we define the salience of an APT adapter as the summation of the parameter salience scores in $W_B$, denoted as $\mathcal{I}(H_{\text{apt}}) = \sum_{i, j}S({W_B}_{i, j})$, to represent each tuning {\ourarchabbr}'s importance\footnote{The salience scores calculated using $W_B$ and $W_A$ are equal, so using either of them will get the same result.}. Given the calculated $\mathcal{I}(H_{\text{apt}})$ for each {\ourarchabbr}, we can then decide where to add new tuning parameters to efficiently improve the pruned {\lmabbr}'s task accuracy.

\noindent \textbf{Dynamically adding APT adapter parameters to recover task performance.} 
With the importance of {\ourarchabbr}s $\mathcal{I}(H_{\text{apt}})$ calculated, the next step of adaptive tuning is to add tuning parameters by increasing the salient tuning layers' ranks $r_{\text{apt}} \in \mathcal{R}_t$ following budget $\Delta_t$. Therefore, firstly, we sort all tuning layers according to their importance score $\mathcal{I}(H_{\text{apt}})$ and linearly increase the ranks of the top-half salient ones. More specifically, when increasing the tuning parameter from $\Delta_t$ to $\Delta_{t'}$, the salient layer's rank is changed from $r_{\text{apt}}$ to $r_{\text{apt}}' = \lfloor r_{\text{apt}} \cdot \frac{\Delta_{t'}}{\Delta_t} \rfloor$ where $\lfloor \cdot \rfloor$ denotes the floor operation. For training stability, when adding parameters and converting $W_B \in \mathbb{R}^{d_o \times r_{\text{apt}}}, W_A \in \mathbb{R}^{r_{\text{apt}} \times d_i}$ to $W_B' \in \mathbb{R}^{d_o \times r_{\text{apt}}'}, W_A' \in \mathbb{R}^{r_{\text{apt}}' \times d_i}$, we concatenate random Gaussian initialized parameters $\mathcal{N}(0, \sigma^2)$ in $W_A$ and zeros in $W_B$ same as the LoRA initialization, so the layer's output remains unchanged before and after new parameters added.




\subsection{Efficient Self-Knowledge Distillation} \label{sec:momentum}
As shown in \cref{tab:ablate-paramalloc}, training pruned {\lmabbr} without knowledge distillation causes significant end-task performance drops. Therefore, we use knowledge distillation in {\ourmethod} to recover the pruned {\lmabbr}'s performance. Still, existing strategies require a fully trained teacher model being put into the GPU with the student during distillation, causing high training time and memory. To avoid extra training costs, we keep duplicating the tuning student layers as teachers during fine-tuning to reduce total training time. Meanwhile, frozen parameters are shared between the student and teacher model during training to reduce memory consumption. We edit the distillation objective in CoFi~\citep{xia_structured_2022} as
\begin{equation}
    \begin{split}
        \mathcal{L} &= \mu \mathcal{L}_{distill} + (1 - \mu) \mathcal{L}_{ft} \\
        \mathcal{L}_{layer} &= \sum_{i=1}^\mathcal{T} \textrm{MSE}(\text{Tr}(H_s^{\phi(i)}), H_t^i)
    \end{split}
\end{equation}
where $\mu$ is a moving term linearly scales from 0 to 1 during distillation to encourage the pre-pruned model vastly fit to the training data, $\mathcal{L}_{distill}$ is the distillation objective from CoFi, and $\mathcal{L}_{ft}$ is the supervised fine-tuning objective. $\mathcal{T}$ is block-wise randomly sampled teacher layers following~\citep{haidar2022rail}, $\phi(\cdot)$ is the teacher-student layer-mapping function that matches the teacher layer to its closest, non-pruned student layer. $\text{Tr}$ denotes the tunable LoRA layer for layer transformation, initialized as an identical matrix $\mathcal{I}$. More implementation details of our self-distillation technique is introduced in \cref{appendix:hyper-param}.


%% file: sections/experiment.tex
\section{Experiments}


\input{tables/merged_main_results_time}

\input{tables/llama_7b_results_time}

To evaluate the training and inference efficiency gains of {\ourmethod}, we compare it with the combined use of PEFT with pruning and distillation baselines. We first describe the natural language understanding and generation tasks targeting different {\lmabbr} backbones, then the setup of baselines and {\ourmethod}. We then report task performance, speed, and memory usage for training and inference costs. 

\subsection{Tasks}

We apply {\ourmethod} to BERT~\citep{devlin-etal-2019-bert}, RoBERTa~\citep{liu2019roberta},  T5\citep{raffel2020exploring}\footnote{For fair comparisons, we use the t5-lm-adapt model, which is only pre-trained on the C4 corpus to make sure the initial {\lmabbr} does not observe downstream tasks in pre-training.}, and LLaMA~\citep{touvron2023llama}. 
For BERT, RoBERTa, and T5 models, we train and evaluate on SST2 and MNLI datasets from the GLUE benchmark~\citep{wang2018glue} and report the dev set accuracy. We also train and evaluate $\text{RoBERTa}_{\text{base}}$ on SQuAD v2.0~\citep{rajpurkar-etal-2018-know} and report the dev set F1 score. For T5 models, we also fine-tune them on CNN/DM~\citep{Nallapati2016AbstractiveTS} and report the ROUGE 1/2/L scores. Meanwhile, We use the GPT-4 generated Alpaca dataset~\citep{alpaca} to fine-tune large LLaMA models and evaluate them with the lm-eval-harness package~\citep{eval-harness} on four tasks from the Open LLM Leaderboard, namely 25-shot ARC~\citep{clark2018think}, 10-shot HellaSwag~\citep{zellers-etal-2019-hellaswag}, 5-shot MMLU~\citep{hendrycks2021measuring}, and zero-shot TruthfulQA~\citep{lin-etal-2022-truthfulqa}.

\subsection{Baselines}
We validate the efficiency benefits of {\ourmethod} for both training and inference by comparing with PEFT, pruning, and distillation methods, along with their combinations. 

\noindent \textbf{LoRA+Prune}: a post-training pruning method over on LoRA-tuned {\lmabbr}s. We use Mask Tuning~\citep{kwon_fast_2022}, a state-of-the-art post-training structured pruning method based on fisher information. Due to that post-training pruning performs poorly on high-sparsity settings, we retrain the pruned {\lmabbr} after pruning to recover its performance.


\noindent \textbf{Prune+Distill}: knowledge distillation has been proved to be a key technique in recovering pruned {\lmabbr}s' task accuracy. In particular, we use the state-of-the-art pruning plus distillation method called CoFi~\citep{xia_structured_2022} which uses $L_0$ regularization for pruning plus dynamic layer-wise distillation objectives. We only compare {\ourmethod} to CoFi with RoBERTa models since the training memory usage of CoFi is too high for larger {\lmabbr}s.

\noindent \textbf{LoRA+Prune+Distill}: to reduce the training memory consumption in pruning and distillation, a simple baseline is to conduct CoFi pruning and distillation but with LoRA parameters tuned only. More specifically, only the $L_0$ module and LoRA parameters are tunable under this setting.  

\noindent \textbf{LLMPruner}~\citep{ma2023llm}: LLMPruner is the state-of-the-art task-agnostic pruning method on LLaMA that prunes its blocks or channels based on salience metrics while using LoRA for fast performance recovery. We compare {\ourmethod} to LLMPruner with fine-tuning on the same GPT-4 generated Alpaca data for fair comparisons.

We also compare {\ourmethod} to PST~\citep{ijcai2022p586} and LRP~\citep{zhang2023pruning}, which are the state-of-the-art parameter-efficient unstructured and structured pruning methods on $\text{BERT}$ model. We leave these results in \cref{sec:appendix-additional-exp}.


\subsection{Evaluation Metrics}
We evaluate {\ourmethod} and baselines on training and inference efficiency, measured in runtime memory and time consumption as follows:

\noindent\textbf{Training Efficiency Metrics}: we report relative training peak memory (Train. Mem.) and relative training speed measured by time to accuracy (TTA\footnote{For instance, 97\% TTA denotes the time spent reaching 97\% of the fully fine-tuned model's performance})~\citep{10.1145/3352020.3352024} compared to full finetuning. For fair comparisons, we consider the training time of the teacher model plus the student for methods using knowledge distillation.

\noindent\textbf{Inference Efficiency Metrics}: we report the inference peak memory (Inf. Mem.) and the relative speedup (Inf. Speed) based on throughput (data processed per second) for inference efficiency. 

Both training and evaluation are conducted on a single A100 GPU. The inference test batch size is 128 for small models while 32 and 4 for LLaMA 7B and 13B models, respectively. We demonstrate detailed training and evaluation setups/implementations in \cref{appendix:hyper-param}.

\subsection{Main Results} \label{sec:main-result}


\noindent \textbf{Overview} We demonstrate the end-task performance of {\ourmethod} comparing to fine-tuning (FT), LoRA-tuning (LoRA), and pruning baselines in \cref{tab:small-model-results} and \cref{tab:llama-7b-results}. Overall, up to 99\% of fine-tuned {\lmabbr}'s task accuracy is maintained when pruning RoBERTa and T5 models leaving 40\% parameters, with only about 70\% training memory consumption than fine-tuning. When pruning LLaMA2-7B models with 70\% parameters remaining, {\ourmethod} recovers 86.4\% task performance on average, together with only 75.8\% training memory usage than LoRA-tuning. Furthermore, {\ourmethod} also significantly reduces end-task performance and training costs compared to the pruning and distillation baselines. The detailed comparisons are shown as follows.

\noindent \textbf{{\ourmethod} speeds up RoBERTa and T5 training 8$\times$ and reduces training memory costs to 30\% in LLaMA pruning compared to LoRA+Prune baseline.} 
Shown in \cref{tab:small-model-results}, when pruning RoBERTa models to 60\% sparsity, {\ourmethod} converges $8.4\times$ faster than the LoRA+Prune baseline with consuming similar GPU memory. {\ourmethod} also prunes T5 models 8.2$\times$ faster than the LoRA+Prune baseline. The reason is that {\ourmethod} adaptively prunes task-irrelevant parameters during training, reducing memory and per-step training time. Adding parameters in salient tuning layers also accelerates {\lmabbr} convergence. 
Also, {\ourmethod} costs less than 24GB of memory when pruning 30\% parameters in LLaMA2-7B models before tuning, which can be easily adapted to the consumer-level GPUs. In contrast, LLM-Pruner costs about 80GB memory when pruning the LLaMA 7B model\footnote{\url{https://github.com/horseee/LLM-Pruner/issues/4}}.

\noindent \textbf{{\ourmethod} achieves 2.5\%-9.9\% higher task performance than the LoRA+Prune baseline with the same pruning sparsities.} 
Presented in \cref{tab:small-model-results} and \cref{tab:llama-7b-results}
, when RoBERTa, T5, and LLaMA models, regardless of size, {\ourmethod} consistently reach higher task performance than the LoRA+Prune. 
With similar inference speedup and memory when pruning RoBERTa models, {\ourmethod} reaches 2.5\% more end-task performance on average.
When pruning T5 models under the 60\% sparsity, the task performance achieved by {\ourmethod} is 5.1\% better than the LoRA+Prune baseline. However, the inference efficiency reached by {\ourmethod} (1.3$\times$ speedup and 81.5\% memory cost) is worse than the LoRA+Prune baseline (2.1$\times$ speedup and 73.4\% memory cost). This is because {\ourmethod} can adaptively prune more decoder parameters, which are also computationally cheaper than encoder parameters (due to shorter output sequence length) but relatively useless for classification tasks. 
For LLaMA2-7B model pruning with 70\% sparsity, {\ourmethod} outperforms LLMPruner with 16.5\% and the LoRA+Prune baseline with 9.9\%, where the inference efficiency improvements of {\ourmethod} is slightly better than both LoRA+Prune and LLMPruner baselines. 


\noindent \textbf{{\ourmethod} reaches on-par performance with the Prune+Distill baseline given the same pruning sparsity but trains 2.5$\times$ faster and costs only 41.6\% memory.}
Compared to the Prune+Distill baseline, {\ourmethod} results in comparable task accuracy (0.9 point drop in MNLI and same in SST2). At the same time, with similar inference efficiency achieved, {\ourmethod} costs only 41.6\% training memory and converges 2.5$\times$ than the Prune+Distill baseline. This is because of the self-distillation technique in {\ourmethod} where no separated teacher model is required in pruning {\lmabbr}s. Moreover, {\ourmethod} achieves better task performance than the LoRA+Prune+Distill baseline as well, with less training time and memory consumption. These results demonstrate that {\ourmethod} successfully tackles the problem where simply combining PEFT and pruning hurts pruned {\lmabbr}'s task accuracy and training efficiency.

\begin{figure}[ht]
    \centering
    \includegraphics[width=0.8\columnwidth]{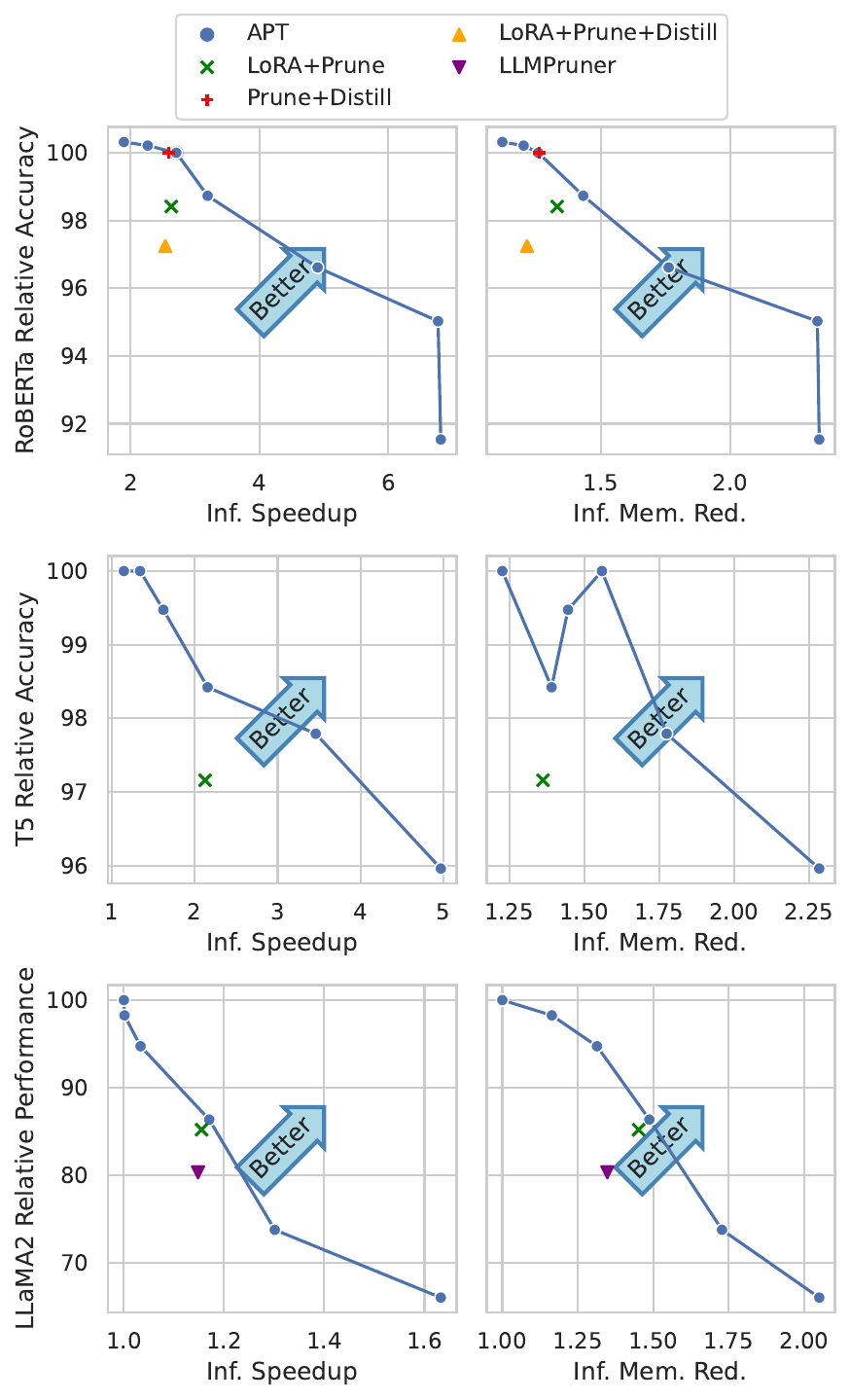}
    \caption{Task performance v.s. relative inference efficiency on RoBERTa, T5, and LLaMA-2 7B models with {\ourmethod} and baselines.}
    \label{fig:prune-tradeoff}
    \vspace{-10pt}
\end{figure}

\subsection{Pruning Sparsity Analysis} \label{sec:sparsity-analysis}

We further show the task performance changing trajectory with different pruning sparsities in \cref{fig:prune-tradeoff}. {\ourmethod} achieves superior inference speedup with less inference memory consumption than baselines targeting the same task performance. Compared to the LoRA+Prune baseline, when pruning RoBERTa models targeting similar task accuracy, {\ourmethod} is 21.8\% faster in inference and is 7\% more memory-efficient. For T5 model pruning with 97\% of dense model performance, {\ourmethod} results in 62.7\% more inference speedup with 24.8\% more inference memory reduction compared to the LoRA+Prune baseline.
When pruning large LLaMA2-7B models, {\ourmethod} speedup is 6.7\% more and reduces 9.2\% more inference memory than the LoRA+Prune baseline, maintaining over 85\% task performance of the dense model.

\subsection{Ablation Study} \label{sec:ablation}
We evaluate the impact of different components in {\ourmethod} by removing the adaptive pruning ($\mathcal{A}_{\text{P}}$), adaptive tuning ($\mathcal{A}_{\text{T}}$), and self-distillation ($\mathcal{D}_{\text{S}}$). Besides end-task performance, we also report the training efficiency metrics for each ablation. 

\noindent \textbf{Adaptive pruning ($\mathcal{A}_{\text{P}}$)} We demonstrate the ablation of adaptive pruning (w/o $\mathcal{A}_{\text{P}}$) for RoBERTa models in \cref{tab:ablate-paramalloc} and LLaMA models in \cref{tab:llama-ablation}. In these cases, we only train {\lmabbr}s with adaptive tuning strategies with supervised fine-tuning objectives without distillation. In such settings, {\ourmethod} w/o $\mathcal{A}_{\text{P}}$ can be recognized as a PEFT method with tuning parameters' sizes adaptively changing during fine-tuning. Hence, the inference efficiency of the trained {\lmabbr}s are the same as full fine-tuning and LoRA. Without pruning, the task performance of RoBERTa reaches 94.4 for SST2 and 87.5 for MNLI (99.8\% fine-tuned {\lmabbr} performance on average). The average performance of the LLaMA model also achieves 96.6\% to its LoRA-tuned counterpart. In addition, we surprisingly find that the RoBERTA training speed with {\ourmethod} w/o $\mathcal{A}_{\text{P}}$ is even 21\% faster than full fine-tuning while costing only 62.2\% memory.
In the meantime, the training memory cost of {\ourmethod} w/o $\mathcal{A}_{\text{P}}$ in LLaMA tuning is higher than LoRA. The reason is that the tuning parameter number of {\ourmethod} will grow larger than static LoRA-tuning. This ablation demonstrates that adaptive pruning is essential in reducing the training memory consumption of LLaMA model fine-tuning, besides benefiting model inference efficiency.

\noindent \textbf{Adaptive tuning ($\mathcal{A}_{\text{T}}$)} In \cref{tab:ablate-paramalloc}, we show results of ablating adaptive tuning (w/o $\mathcal{A}_{\text{T}}$) where the tuning parameters are static when pruning RoBERTa models.
Without $\mathcal{A}_{\text{T}}$, the model's performance decreases to 93.2/84.4, leading to a similar performance as the LoRA+Prune baseline (93.0/84.0). Moreover, equally increasing parameters across all layers instead of adding parameters based on salience notably hurts the task accuracy (84.4 on MNLI compared to 86.4). At the same time, $\mathcal{A}_{\text{T}}$ helps the model converge 16\% faster than static LoRA training.
For ablation results in LLaMA models shown in \cref{tab:llama-ablation}, we observe that $\mathcal{A}_{\text{T}}$ recovers the model performance under 50\% pruning setting (38.2 compared to 35.8). However, the difference under 70\% pruning is insignificant. Meanwhile, if calculating the pruning parameter salience without using kurtosis to consider outliers parameters, the pruned {\lmabbr}'s performance substantially drops from 50.0 to 38.1. We conclude that $\mathcal{A}_{\text{T}}$ substantially improves {\lmabbr} training speed and end-task performance. For large LLaMA-based {\lmabbr} pruning, and outlier parameters are essential to recovering the pruned large LLaMA-based models' capabilities.

\input{tables/ablate_elastic_time}

\input{tables/llama_7b_ablation}



\noindent \textbf{Self-distillation ($\mathcal{D}_{\text{S}}$)}
Shown in \cref{tab:ablate-paramalloc}, tuning {\ourmethod} adapters dynamically without distillation objectives gets 1.35 worse task accuracy on average. However, pruning RoBERTa models without self-distillation is 22.5\% faster and costs 11.7\% less training memory. This result indicates the effectiveness of leveraging knowledge distillation to recover pruned {\lmabbr} performance, but conducting distillation will result in extra training costs regarding both time and memory. Detailed comparisons of self-distillation and traditional, static distillation strategies are shown in \cref{appendix:distillation}.

Besides the ablation study results demonstrated above, we show the detailed analysis of adaptive pruning and tuning's effect on {\lmabbr}s' end-task performance, training, and inference efficiency in \cref{sec:analysis}.


\section{Limitation and Discussion} \label{sec:discussion}
\noindent \textbf{Towards better performance gain and inference speedup of large {\lmabbr} in limited resource settings.}
By comparing \cref{tab:small-model-results} to \cref{tab:llama-7b-results}, we notice the performance gap in pruned LLaMA models is larger than smaller {\lmabbr}s because we use distillation-free settings in large {\lmabbr} pruning to reduce training memory consumption. One can improve performance-efficiency trade-offs with better memory-efficient distillation, parameter sharing, and re-allocation strategies. Furthermore, because of the hardware features of Ampere-architecture GPUs, layer dimensions divisible by 8 for FP16 and divisible by 16 for Int8 would reach more realistic speedups. One possible direction is to explore a higher level of structured pruning, for example, grouped neurons and dimensions, in LLMs.

\noindent \textbf{Training could be unstable because of parameter shape changes.} Since we adjust tuning parameters dynamically during training, newly initialized parameters are added to the model while existing parameters are pruned. We reset the optimizer every time after each parameter size changes to avoid stability issues, but this strategy might cause unstable training. Meanwhile, the time of selecting the teacher checkpoints during training highly affects the pruned model's performance, whereas non-converged or sparse teachers do not help in performance recovery. The pruned {\lmabbr}s' end-task accuracy could benefit from better and more stable strategies in adaptive pruning and tuning.

\noindent \textbf{Could non-linear adapters perform better for performance recovery?} To avoid inference time and memory overhead, we specifically adapt {\ourarchabbr} to LoRA since the added tuning parameters can be merged after LMs' training. However, low-rank decomposition does not add more complexity to a {\lmabbr}, whereas the model's overall representation capacity doesn't increase. The adaptation with a wider range of adapters, such as Prefix-tuning~\citep{li2021prefix}, HAdapters~\citep{houlsby2019parameter}, and Parallel-adapters~\citep{he_towards_2022}, could be better explored.

%% file: tables/merged_main_results_time.tex
\begin{table*}[t!]
\resizebox{1.0\linewidth}{!}{
\begin{tabular}{@{}ll|rrrr|rrrr@{}}
\toprule
Model                                           & Method         & MNLI          & SST2          & SQuAD v2      & CNN/DM                  & Train Time($\Downarrow$) & Train Mem($\Downarrow$) & Inf Time($\Downarrow$) & Inf Mem($\Downarrow$) \\ \midrule
\multirow{5}{*}{$\text{RoBERTa}_{\text{base}}$} & FT             & 87.6          & 94.8          & 82.9          & -                       & 100.0\%      & 100.0\%     & 100.0\%    & 100.0\%   \\
                                                & LoRA           & 87.5          & 95.1          & 83.0          & -                       & 2137.0\%        & 60.5\%      & 100.0\%    & 100.0\%   \\ \cmidrule(l){2-10} 
                                                & LoRA+Prune   & 84.0          & 93.0          & 79.2          & -                       & 5128.3\%        & \textbf{60.5\%}      & \textbf{38.0\%}    & \textbf{75.1\%}    \\
                                                & Prune+Distill   & \textbf{87.3}          & \textbf{94.5}          &   -        & -                       & 1495.3\%        & 168.5\%      & 38.6\%    & 79.2\%    \\
                                                & LoRA+Prune+Distill & 84.2          & 91.9          & -             & -                       & 6534.6\%        & 141.4\%     & 39.4\%    & 82.3\%    \\  \cmidrule(l){2-10} 
                                                & \ourmethod     & 86.4 & \textbf{94.5} & \textbf{81.8} & -                       & \textbf{592.1\%}       & 70.1\%      & 41.3\%    & 78.1\%    \\ \midrule
\multirow{4}{*}{$\text{T5}_{\text{base}}$}      & FT             & 87.1          & 95.2          & -             & 42.1/20.3/39.4          & 100.0\%      & 100.0\%     & 100.0\%    & 100.0\%   \\
                                                & LoRA           & 87.0          & 95.0          & -             & 38.7/17.2/36.0          & 255.5\%       & 62.0\%      & 100.0\%    & 100.0\%   \\ \cmidrule(l){2-10} 
                                                & LoRA+Prune   & 80.9          & 92.3          & -             & 36.7/15.7/33.9          & 4523.5\%        & \textbf{62.0\%}      & \textbf{47.1\%}    & \textbf{73.4\%}    \\ \cmidrule(l){2-10} 
                                                & \ourmethod     & \textbf{87.0} & \textbf{95.0} & -             & \textbf{38.6/17.0/35.8} & \textbf{484.7\%}       & 73.9\%      & 74.6\%    & 81.5\%    \\ \bottomrule
\end{tabular}
}
\caption{RoBERTa and T5 pruning with {\ourmethod} compared to baselines under 60\% sparsity. We measure the training and inference efficiency with {\lmabbr}s pruned on the SST2 task. Training speed is measured via 97\% accuracy TTA. All efficiency metrics are normalized to FT. $\Downarrow$ denotes smaller is better. The best-pruned results are \textbf{bold}. Raw efficiency results are reported in \cref{tab:raw-efficiency}.}
\label{tab:small-model-results}
\vspace{-5pt}
\end{table*}

%% file: tables/llama_7b_results_time.tex
\begin{table*}[htbp]
\centering
\resizebox{1.0\linewidth}{!}{
\begin{tabular}{@{}l|rrrrr|rrrr@{}}
\toprule
Method       & ARC                            & HellaSwag                      & MMLU                           & TruthfulQA                     & Avg.                           & Train Time($\Downarrow$) & Train Mem ($\Downarrow$) & Inf Time($\Downarrow$) & Inf Mem($\Downarrow$) \\ \midrule
LLaMA 2 7B   & 53.1                           & 77.7                           & 43.8                           & 39.0                           & 53.4                           & -            & -           & -          & -         \\ 
LoRA         & 55.6                           & 79.3                           & 46.9                           & 49.9                           & 57.9                           & 100.0\%      & 100.0\%     & 100.0\%    & 100.0\%   \\ \midrule
LoRA+Prune & \textbf{46.8} & 65.2                           & 23.9                           & 46.2                           & 45.5                           & 180.9\%       & 100.0\%     & 115.5\%    & 68.9\%    \\
LLMPruner    & 39.2                           & 67.0                           & 24.9                           & 40.6                           & 42.9                           & \textbf{86.9\%}      & 253.6\%     & \textbf{114.8\%}    & 74.2\%    \\ \midrule
APT          & 45.4                           & \textbf{71.1} & \textbf{36.9} & \textbf{46.6} & \textbf{50.0} & 106.0\%       & \textbf{75.8\%}      & 117.0\%    & \textbf{67.2\%}    \\ \bottomrule
\end{tabular}
}
\caption{
    LLaMA 2 7B 30\% sparsity pruning results with GPT4-generated Alpaca dataset, evaluated on the Open LLM leaderboard few-shot tasks. Training speed is measured via training time per step. We do not compare to distillation baselines because the training cost of distillation is too large, and we also compare {\ourmethod} to LLMPruner since it is dedicated to large {\lmabbr} pruning. All efficiency metrics are normalized to LoRA. $\Downarrow$ denotes smaller is better. The best-pruned results are \textbf{bold}. Raw efficiency results are reported in \cref{tab:raw-llama-efficiency}.}
\label{tab:llama-7b-results}
\vspace{-15pt}
\end{table*}

%% file: tables/ablate_elastic_time.tex
\begin{table}[htbp]
\resizebox{1.0\linewidth}{!}{
\begin{tabular}{@{}l|rr|rr@{}}
\toprule
Method                                    & SST2 & MNLI & Train Time($\Downarrow$) & Train Mem($\Downarrow$) \\ \midrule
\ourmethod                                & \textbf{94.5} & 86.4 & 592.1\%       & 70.1\%        \\
\tableindent w/o $\mathcal{A}_{\text{P}}$ & 94.4 & \textbf{87.5} & \textbf{82.6\%}      & 62.2\%        \\
\tableindent w/o salience                 & 94.3 & 84.7 & 609.8\%       & 65.0\%        \\
\tableindent w/o $\mathcal{A}_{\text{T}}$ & 93.2 & 84.5 & 684.9\%       & 64.4\%        \\
\tableindent w/o $\mathcal{D}_{\text{S}}$ & 92.9 & 85.3 & 483.1\%       & \textbf{61.9\%}        \\ \bottomrule
\end{tabular}
}
\caption{Results of ablating salience-based allocation strategy and {\ourarch} with RoBERTa-base model, with relative training efficiency metrics to fine-tuning.}
\label{tab:ablate-paramalloc}
\vspace{-5pt}
\end{table}

%% file: tables/llama_7b_ablation.tex
\begin{table}[htbp]
\centering
\resizebox{1.0\linewidth}{!}{
\begin{tabular}{@{}lrr|rrrrr@{}}
\toprule
                            & Sparsity & T.M.   & ARC  & HellaSwag & MMLU & TruthfulQA & Avg. \\ \midrule
APT                         & 30\%    & 75.8\% & 45.4 & 71.1      & 36.9 & 46.6       & 50.0 \\ \midrule
\tableindent w/o $\mathcal{A}_{\text{P}}$     & 100\%    & 102.4\% & 53.8 & 79.1      & 46.9 & 48.4       & 57.1 \\
\tableindent w/o kurtosis   & 30\%    & 75.9\% & 47.2 & 39.7    & 23.0 & 42.3       & 38.1 \\
\tableindent w/o $\mathcal{A}_{\text{T}}$ & 30\%    & 76.1\% & 44.2 & 70.1      & 40.8 & 45.1       & 50.0 \\ \midrule
APT                         & 50\%    & 60.2\% & 29.8 & 48.9      & 26.7 & 47.6       & 38.2 \\
\tableindent w/o $\mathcal{A}_{\text{T}}$ & 50\%    & 60.1\% & 27.9 & 46.2      & 24.5 & 44.7       & 35.8 \\ \bottomrule
\end{tabular}
}
\caption{LLaMA 2 7B model ablation results under 30\% and 50\% sparsity settings. T.M. denotes relative training memory compare to LoRA-tuning.}
\label{tab:llama-ablation}
\vspace{-5pt}
\end{table}

%% file: sections/conclusion.tex
\section{Conclusion}
We design {\ourmethod} to adaptively identify {\lmabbr}s' pruning and tuning parameters during fine-tuning, improving both training and inference efficiency. {\ourmethod} prunes small {\lmabbr}s faster while pruning large {\lmabbr}s with less memory consumption. With using similar memory costs as LoRA, {\ourmethod} prunes small {\lmabbr}s 8$\times$ faster than the LoRA plus pruning baseline. In large {\lmabbr} pruning, {\ourmethod} maintains 87\% performance with only 30\% pruning memory usage when 70\% {\lmabbr} parameter retained. {\ourmethod} opens new directions to pruning {\lmabbr}s in fine-tuning for resource-limited settings, allowing wider usage of {\lmabbr}s in practical applications. In the future, we could adapt {\ourmethod} to more PEFT architectures and target better performance-efficiency trade-offs for billion-level large {\lmabbr}s. Meanwhile, we hope future research will continue to find efficient and accurate techniques to identify salient structures in {\lmabbr}s based on our formulated setting.

%% file: sections/impact_statement.tex
\section*{Impact Statement}
This paper introduces APT, a paradigm for improving the efficiency of training and inference in pre-trained LMs. While our primary goal is to advance machine learning, particularly in the efficiency of LMs and their applications, we recognize potential broader societal impacts. APT significantly reduces training and inference costs and contributes to lower resource consumption for a wide range of applications. This could have a positive environmental impact but might lead to potential model misuse due to lower resource requirements. Additionally, while APT does not introduce new ethical concerns, it might inherit existing issues in language models, for example, biases in training data. We explicitly ask users of APT to be aware of these risks and follow best practices in data selection and model monitoring to mitigate potential harms.

%% file: sections/appendix.tex
\section{Hyperparameter and Training Details} \label{appendix:hyper-param}
Our hyper-parameter settings are shown in \cref{tab:appendix-hyperparam}. For GLUE task fine-tuning, we follow the hyper-parameter setting of CoFi~\citep{xia_structured_2022}, separating the tasks into big (MNLI, SST2, QNLI, QQP) and small (MRPC, CoLA, RTE, STSB) based on the dataset size. For instruction tuning on the Alpaca dataset, we train the pruned model for 15 epochs after the pre-tuning pruning process to make sure they converge. However, in practice, such training epochs can be reduced. To adaptively increase the tuning parameters in the {\lmabbr}, at the start of fine-tuning, we initialize adapter ranks to 8, with salient layers' ranks linearly increased. The scaling factors are set as 2 statically. Since evaluating billion-level LLaMA models during instruction tuning with all evaluation tasks would be time-consuming, we did not do the TTA evaluation as small models. We do not conduct any hyper-parameters search for any training for fair comparison.

\input{tables/appendix_hyperparam}

When pruning {\lmabbr}s with {\ourmethod}, following \citep{xia_structured_2022}, we first prune and train the {\lmabbr} with the self-distillation objective, and then fine-tune the pruned {\lmabbr} to recover its end-task performance. Given $T$ pruning training steps in total, we set a pre-determined target sparsity $\gamma_T$ (defined as the ratio of pruned parameter size to the total parameter size) and use cubic scheduling to control the {\lmabbr} parameter size, where $\gamma_t = \gamma_T + (1-\gamma_T) (1 - \frac{t}{T})^3$. We adaptively increase the tuning parameters in the pruning stage but restrict them to a specific limit $\Delta_t$ at each training step $t$.
Towards better training stability in {\lmabbr} pruning, we gradually decrease the pruning masks of pruned blocks by $\alpha < 1$ instead of instantly setting them from ones to zeros. We also use the exponential moving-averaged salience in~\citep{zhang2023adaptive} when calculating the salience score during fine-tuning.

\section{Block salience calculation and correlations} \label{appendix:salience-unify}
As addressed in \cref{sec:apt}, we use the compressed weight-gradient production as the salience metric for identifying the tuning and pruning parameter blocks in {\lmabbr}s. Previous works~\citep{sanh_movement_2020} use salience score defined as the magnitude of the parameters' weight-gradient production, where given a linear layer $H = WX$ (we omit the bias term here for simplicity) in model parameters $\Theta$ trained on the objective $\mathcal{L}$, the salience scoring function $S$ is defined as:

\begin{equation} \label{eq:sensitivity}
    \begin{split}
        S(W_{i,j}) &= \sum_{(x, y) \in \mathcal{D}} s(W_{i,j}, x, y) \\
        &= \sum_{(x, y) \in \mathcal{D}}|\frac{\partial \mathcal{L}(x, y | \Theta)}{\partial W_{i,j}} \cdot W_{i,j}| \\
        S(W_{:,j}) &= \sum_{(x, y) \in \mathcal{D}}\sum_{i}|\frac{\partial \mathcal{L}(x, y | \Theta)}{\partial W_{i,j}} \cdot W_{i,j}| \\
        &= \sum_{(x, y) \in \mathcal{D}} (\sum_{i}|\frac{\partial \mathcal{L}(x, y | \Theta)}{\partial X_{j, i}} \cdot X_{j, i}|)
    \end{split}
\end{equation}

where $x, y$ are the inputs and labels sampled from the training batch $\mathcal{D}$. $S(W_{i,j})$ denotes the unstructured, sparse parameter's salience, and $S(W_{:, j})$ denotes the salience score of a block in the weight $W$ (for example, rows, columns, attention heads, etc.).

When applying this equation to {\ourarchabbr} layers as defined in \cref{eq:elastic-lora}, there are three different consistent dimensions, namely input dimension $j$, output dimension $i$, and tuning rank dimension $k$. Therefore, the combined salience (including tuning low-rank weights and the frozen weight) of the parameter block shall be calculated as follows:
\begin{equation} \label{eq:appendix-elastic-salience}
    \begin{split}
        S(H, i) &= \sum_l \frac{\partial \mathcal{L}(x, y | \Theta)}{\partial H(X)_{i,l}} \cdot H(X)_{i,l} \\
                &= \sum_p \frac{\partial \mathcal{L}(x, y | \Theta)}{\partial W_{i,p}} \cdot W_{i,p} \\
                &+ s\cdot \sum_q \frac{\partial \mathcal{L}(x, y | \Theta)}{\partial {W_B}_{i,q}} \cdot {W_B}_{i,q} \\
        S(H, j) &= \sum_l \frac{\partial \mathcal{L}(x, y | \Theta)}{\partial X_{j,l}} \cdot X_{j,l} \\
                &= \sum_p \frac{\partial \mathcal{L}(x, y | \Theta)}{\partial W_{p,j}} \cdot W_{p,j} \\
                &+ s\cdot \sum_q \frac{\partial \mathcal{L}(x, y | \Theta)}{\partial {W_A}_{q,j}} \cdot {W_A}_{q,j} \\
        S(H, k) &= s\cdot \sum_l \frac{\partial \mathcal{L}(x, y | \Theta)}{\partial {W_A}_{k,l}} \cdot {W_A}_{k,l} \\
                &= s\cdot \sum_l \frac{\partial \mathcal{L}(x, y | \Theta)}{\partial {W_B}_{l,k}} \cdot {W_B}_{l,k} \\
    \end{split}
\end{equation}
Therefore, we can notice that the real block-wise salience of the LoRA layer shall be the sum of the block-wise frozen weight salience and the corresponding tuning weight. Hence, the existing work~\citep{zhang2023pruning} that only uses the tuning block salience as layer salience leads to sub-optimal pruning results. Meanwhile, we shall also notice the correlation between the input-, output-dimension, and tuning rank dimensions, which are the summation of the weight-gradient production of parameters on different dimensions. 

\section{Adaptive Pruning and Tuning Details} \label{appendix:binary-search}

\begin{algorithm*}[t!]
 	\caption{Adaptive Pruning and Tuning} 
 	\label{alg:epa}
 	\begin{algorithmic}[1]
 		\STATE {{\bfseries Input:} Model $f$; Training dataset $ \mathcal{D} $; total training steps $ T $; Adjustment step set $\mathcal{T}$; Training target $\mathcal{L}$; Initial parameters and masks $\Theta_0, M_0$, training memory budget $\Delta$; Parameter number constraint $\gamma$; Hyperparameters $\alpha\,\beta$.}  
 		\FOR{$ t = 1, \dots, T $}  
            \STATE Forward pass: \( L \leftarrow \mathcal{L}(f(\Theta_t, D_t)) \)
            \STATE Cache the batch-sequence summed hidden states: $\widetilde{H} \leftarrow \sum_{i, j} (|H|)_{ij}$
 		\STATE Backward pass: \( \nabla_{\Theta_t} L \leftarrow \frac{\partial \mathcal{L}(f(\Theta_t, D_t))}{\partial \Theta_t} \)
            \STATE Calculate approximated salience: $\widetilde{S}(m_i) \leftarrow \widetilde{H} \cdot \sum_{i, j} (|\nabla_{H} L|)_{ij}$
            \STATE Update global scores: $\overline{S}^{(t)}(m) \leftarrow \beta \overline{S}^{(t-1)}(m) + (1-\beta) \widetilde{S}(m)$;
 		\STATE Select blocks: $M_1, M_0 \leftarrow$ Binary search against constraint \cref{eq:parameter-constraint}, with scores $\overline{S}^{(t)}(m)$;
            \STATE Update masks: $M^{(t)}_1 \leftarrow min(1, M^{(t-1)}_1 + \alpha)$, $M^{(t)}_0 \leftarrow max(0, M^{(t-1)}_0 - \alpha)$;
            \STATE Update parameters: $\Theta_{t+1} \leftarrow \Theta_t - \alpha \nabla_{\Theta_t} L$
 		\ENDFOR
 		\STATE \textbf{Output:}  { Parameters and masks $ \Theta^{(T)}, M^{(T)}$.} 
	\end{algorithmic}
\end{algorithm*}

We show the detailed algorithm description of our Lightweight Parameter Adjustment as described in \cref{sec:apt} in \cref{alg:epa}. For the details of the algorithm, we first sort all blocks by the salience density, defined as the block salience divided by the number of parameters in the block. For instance, given a RoBERTa-base model with the hidden dimension $d_m = 768$, the number of transformer layers $n_L = 12$, the number of attention heads $n_h = 12$, and the number of FFN neurons $n_f = 3072$, we have:
\begin{align}
    \mathcal{C}_{\text{head}} &= 4 \times d_m \times d_m / n_h = 196608\\
    \mathcal{C}_{\text{neuron}} &= 2 \times d_m = 1536\\
    \mathcal{C}_{\text{dimension}} &= n_L \times (4 d_m + 2 n_f) = 110592
\end{align}
We also omit the bias term for density calculation since it takes up less than 1\% of {\lmabbr}'s parameters. Since the number of heads, neurons, and hidden dimensions is ever-changing during pruning, we re-calculate the density after executing each parameter size change. Meanwhile, for T5 and LLaMA-like models, the FFN layers are gated, consisting of up-, gate-, and down-projection linear layers. Therefore, the number of layers in FFN shall be three instead of two in these {\lmabbr}s. Furthermore, for encoder-decoder {\lmabbr}s like T5, the cross-attention layers in the decoder shall also be counted.

After sorting the blocks by salience density, as {\lmabbr}'s parameter size monotonically increases with the number of MHA heads, FFN neurons, and hidden dimensions, we conduct a binary search algorithm to identify the blocks shall be retained as {\lmabbr}'s parameter size monotonically increases with the number of MHA heads, FFN neurons, and hidden dimensions. Specifically, given a sorted list of $N$ blocks $B = \{b_1,b_2,...,b_N\}$ and function $f$ for identifying the block's category where
\begin{equation}
    f(b_i) = 
    \begin{cases} 
    0 & \text{if } b_i \text{ is a head} \\
    1 & \text{if } b_i \text{ is a neuron} \\
    2 & \text{if } b_i \text{ is a dimension} \\
    \end{cases}
\end{equation}
given any index $i$, we can calculate the parameter number of the {\lmabbr} consisting of the top-$i$ blocks by:
\begin{equation}
\begin{split}
    \mathcal{C}_{\text{top-}i} &= (4 d_h' \cdot n_h' + 2 n_f') \cdot d_m' \\
    n_h' &= \sum_{j=0}^{i-1} \delta(0, f(b_j)) \\
    n_f' &= \sum_{j=0}^{i-1} \delta(1, f(b_j)) \\
    d_m' &= \sum_{j=0}^{i-1} \delta(2, f(b_j)) \\
\end{split}
\end{equation}
where $\delta(i, j)$ is the Kronecker delta function that valued 1 if $i=j$ and otherwise 0. Hence, we can use binary search to get the top-$i$ salient blocks, which shall be retained given a parameter constraint, and the rest of the block shall be pruned. In our implementation, for training stability, we do not set the pruned blocks' corresponding masks to 0 directly but gradually decrease their values by $\alpha = 0.01$.


\input{tables/bert_additional_results}

\input{tables/roberta_detailed_results}

\section{Additional Baseline Comparisons} \label{sec:appendix-additional-exp}
In this section, we further compare {\ourmethod} to existing parameter-efficient pruning methods, such as PST and LRP. In the meantime, we also show detailed results of {\ourmethod} pruning compared to the LoRA+Distill baseline with more tasks in the GLUE benchmark and LLaMA-2 13B model pruning results.

\subsection{Comparison to PST and LRP}
We compare {\ourmethod} with the state-of-the-art joint use of unstructured pruning~\citep{ijcai2022p586} and structured pruning~\citep{zhang2023pruning} with PEFT on $\text{BERT}_{\text{base}}$ model, showing in \cref{tab:bert-additional-result}. We can see that {\ourmethod} outperforms existing baselines in both 50\% and 10\% pruning density settings with a notable margin. The performance gain is credited to our more accurate pruning strategy considering frozen and tuning parameters. At the same time, our efficient self-distillation technique used in conjunction with salient parameters added in training also boosts performance recovery.

\subsection{Further Comparison to LoRA+Distill}
We show the detailed comparison between {\ourmethod} and the LoRA+Distill baseline in \cref{tab:roberta-detailed-results}. {\ourmethod} reaches superior task performance compared to the baseline in all seven GLUE tasks listed in the table, with on average 93.5\% fine-tuned {\lmabbr} performance maintained, notably outperforming the joint use of LoRA and knowledge distillation. In particular, the results of STS-B cannot be reproduced when conducting CoFi distillation with LoRA parameters tuned only, so we exclude the comparison on STS-B. Among the other seven tasks in the GLUE benchmark, we find that tasks with relatively smaller dataset sizes, namely CoLA, MRPC, and RTE, reach superior performance gain when using {\ourmethod}. We conclude that this is because, compared to simple fine-tuning, knowledge distillation with salient parameters added in training is more robust and not prone to overfitting the training data.

\subsection{LLaMA-2 13B Pruning Results}
\input{tables/llama_results}

As shown in \cref{tab:llama-results}, when pruning LLaMA-2 13B models, {\ourmethod} maintains 90.0\% performance of the unpruned LoRA-tuned baseline. Compared to the pruning result on 7B models that maintain 86.4\% dense model performance, better accuracies can be recovered in larger models (13B). At the same time, under the same pre-tuning pruning settings, {\ourmethod} performs better than the LLMPruner baseline on all four evaluation tasks, indicating the effectiveness of considering outlier parameters in large {\lmabbr} pruning. Nonetheless, the LoRA+Prune baseline reaches slightly better results than {\ourmethod} when pruning 13B models, illustrating that there is still room for improving pre-tuning pruning methods in future works. More specifically, among the four tasks we use for evaluating large {\lmabbr}s, TruthfulQA benefits the most from Alpaca fine-tuning. We can see that {\ourmethod} reaches superior results on TruthfulQA than existing baselines regardless of model size. The {\lmabbr}'s capabilities on ARC and HellaSawg downgrade the most when pruning large {\lmabbr} before fine-tuning, implying possibilities of catastrophic forgetting in this paradigm.

\section{Efficiency and Performance Tradeoff Analysis}

\begin{figure*}[t!]
    \centering
    \includegraphics[width=\textwidth]{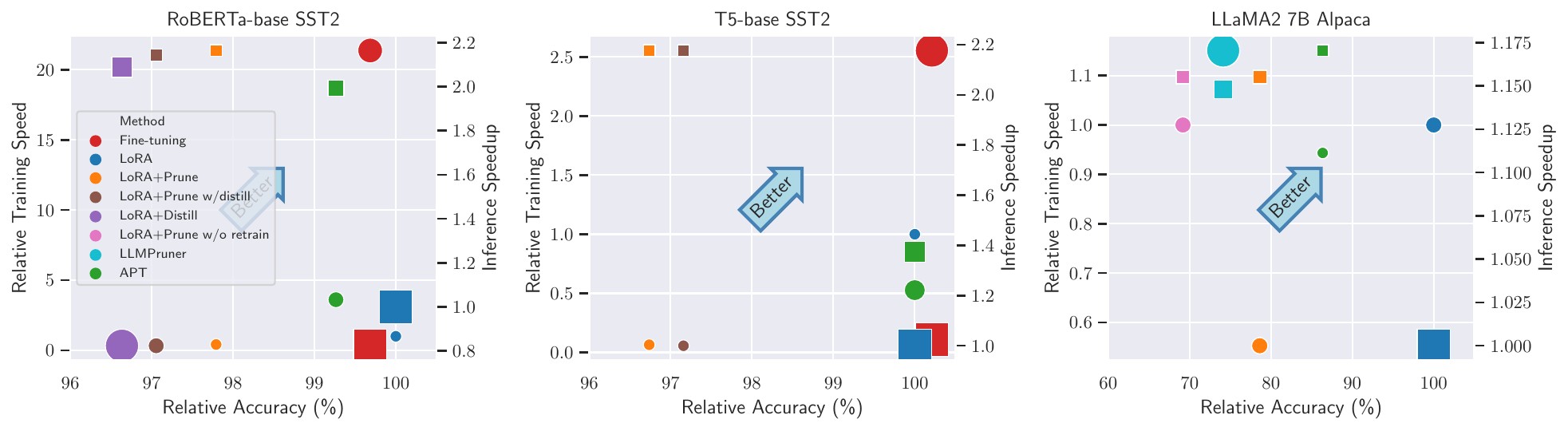}
    \caption{The performance-efficiency tradeoff of {\ourmethod} compared to baseline methods. All metrics are normalized using LoRA tuning w/o pruning as the baseline. The circular dots with vertical axes on the left indicate training speed v.s. performance, with their sizes denoting the peak training memory usage. The squared dots with axes on the right indicate inference speedup v.s. performance, with sizes denoting inference memory usage.}
    \label{fig:perf-efficiency-tradeoff}
    \vspace{-10pt}
\end{figure*}

We use \cref{fig:perf-efficiency-tradeoff} to clearly show the {\lmabbr}s' end-task performance and efficiency tradeoffs between different tuning, pruning, and distillation baselines. We add several extra baselines to conduct more detailed comparisons between {\ourmethod} with existing PEFT, pruning, and distillation methods: 

\noindent \textbf{LoRA+Prune w/distill}: we first use LoRA to fully converge a model on the task dataset, and then use Mask-Tuning~\citep{kwon_fast_2022} to prune the {\lmabbr}. Afterward, we utilize the converged model before pruning as the teacher model and distill its knowledge to the pruned student model with static knowledge distillation objectives.

\noindent \textbf{LoRA+Prune w/o retrain}: we use Mask-Tuning to prune a LoRA-tuned converged model but do not conduct any retraining to recover the pruned models' performance. Therefore, the {\lmabbr}'s training time will be reduced, yet its performance is lower than the LoRA+Prune baseline.

With the same target sparsity in RoBERTa and LLaMA pruning setups, {\ourmethod} achieves on-par end-task performance with full fine-tuning and LoRA tuning baselines. Meanwhile, {\ourmethod}-tuned models reach similar or even better inference time and memory efficiency than existing baselines. {\ourmethod}-pruned T5 {\lmabbr}s' inference efficiency is slightly worse because more decoder parameters (with less computations happening) are pruned than the baselines. Moreover, when pruning RoBERTa and T5 models, {\ourmethod} achieves faster training time than all pruning and distillation baselines. Specifically, the training speed of {\ourmethod} in RoBERTa models is even higher than LoRA tuning without pruning. In LLaMA model pruning, {\ourmethod} costs significantly less training memory than both LLMPruner and LoRA+Prune baselines.

\section{Pruning Sparsity Analysis} \label{appendix:sparsity}

\begin{figure}[ht!]
  \centering
  \begin{minipage}[b]{\textwidth}
    \subfloat[Comparison of different initial ranks of LoRA layers pruning with {\ourmethod} on RoBERTa with SST2 task accuracy, relative training peak memory and speed to 97\% fine-tuning accuracy to the fine-tuning model.]{
    \includegraphics[width=0.45\columnwidth]{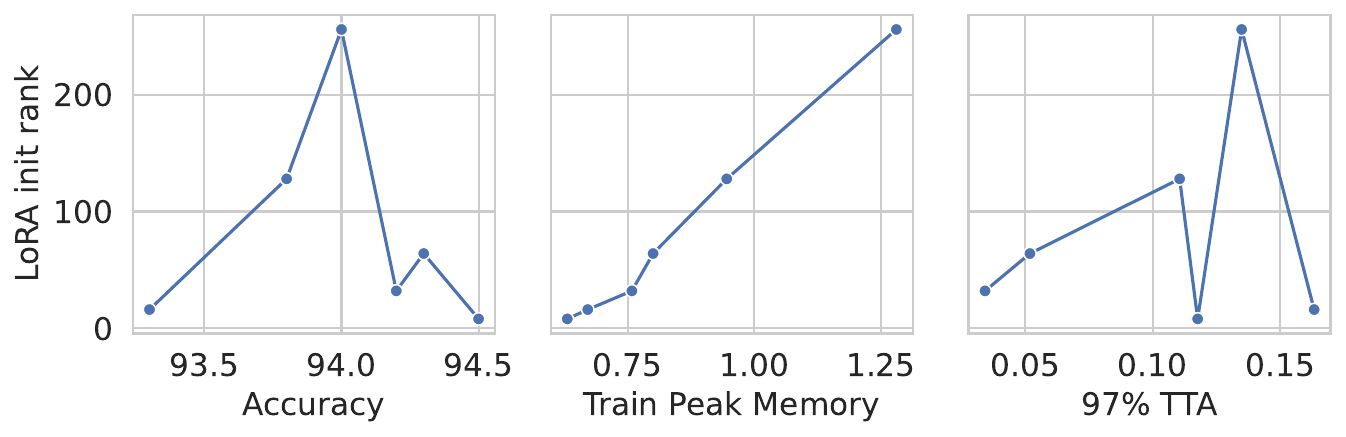}
    \label{fig:rank-tradeoff}
    }
    \hfill
    \subfloat[Training initial sparsity trade-off with 30\% target sparsity model's relative performances to the LoRA-tuned LLaMA2-7B and 13B models.]{
        \includegraphics[width=0.45\columnwidth]{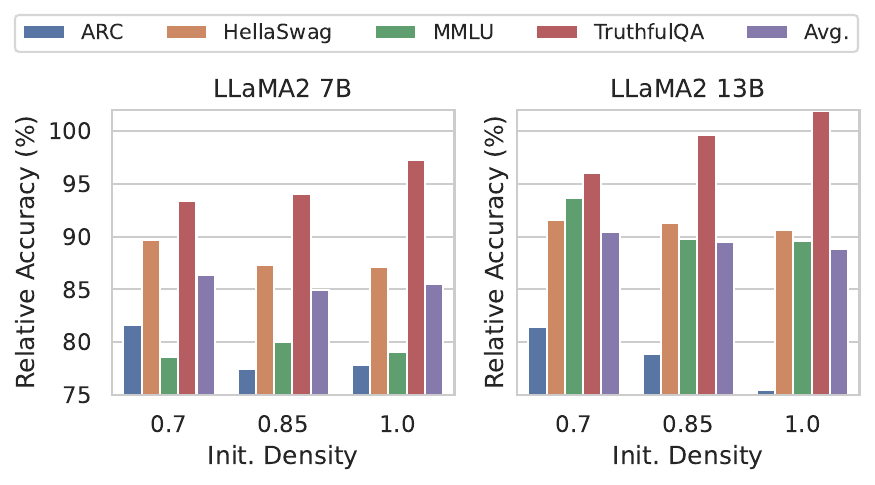}
        \label{fig:llama-init-tradeoff}
    }
  \end{minipage}
  \caption{Detailed analysis in {\ourmethod} with different initial, target sparsities, and adaptive tuning schedules.}
  \label{fig:figureTableCombo}
\end{figure}



We further show the task performance changing trajectory with different pruning sparsities in \cref{fig:prune-tradeoff}. {\ourmethod} achieves superior inference speedup and less inference memory consumption than baselines targeting the same task performance. Compared to the LoRA+Prune baseline, when pruning RoBERTa models targeting similar task accuracy, {\ourmethod} gains 21.8\% more inference speedup and 7\% more memory reduction. For T5 model pruning with 97\% dense model performance maintained, {\ourmethod} results in 62.7\% more inference speedup with 24.8\% more inference memory reduced compared to the LoRA+Prune baseline.
When pruning large LLaMA2-7B models, {\ourmethod} prunes gets 6.7\% more speedup and 9.2\% more inference memory reduction than the LoRA+Prune baseline, with about 85\% dense model performance maintained.

\section{Distillation Strategy Comparison} \label{appendix:distillation}

\input{tables/ablate_distillation}

We show the further analysis in \cref{tab:ablate-distillation} to compare the self-distillation technique we use in {\ourmethod} and traditional knowledge distillation methods. When ablating the dynamic layer mapping strategy in our self-distillation approach, the {\lmabbr} performance decreased by 0.8\% with similar training time and memory consumption. When training without distillation objectives (w/o self-distillation), the {\lmabbr} performance drops by 1.7\%. Nonetheless, the training is slightly faster with less memory costs. These results present that using distillation objectives for better {\lmabbr} task performance will sacrifice training efficiency as a tradeoff.
Furthermore, we also demonstrate the comparisons with existing static knowledge distillation strategies, using the converged full-parameter fine-tuned {\lmabbr} (FT teacher) and LoRA-tuned {\lmabbr} (LoRA teacher) as the teacher model. We calculate the time consumption for both teacher and student training when using these distillation baselines. As shown in \cref{tab:ablate-distillation}, using fully fine-tuned models as the teacher will incur more memory cost than dense model fine-tuning, while {\ourmethod} only consumes 70\%. In the meantime, the training convergence speed of {\ourmethod} training is two times faster than the traditional knowledge distillation method with a fine-tuned teacher. Furthermore, using a LoRA-tuned model as the teacher will result in extremely slow training speed. In addition, simply tuning the LoRA layers with knowledge distillation objectives doesn't help reduce the training memory consumption, as the memory consumption is still 96.1\% than full fine-tuning.

\section{Adaptive Pruning and Tuning Analysis} \label{sec:analysis}


\noindent \textbf{Effects of adaptive tuning strategies on end-task performance and training efficiency.}
As the trajectories shown in \cref{fig:rank-tradeoff}, simply enlarging the initial tuning parameter number in {\ourmethod} will not improve or even hurt the model's final performance. Moreover, the training memory consumption grows even higher than fine-tuning when the tuning layer ranks become extremely large (initial ranks set as 256). Therefore, this result proves that adding tuning parameters according to layer salience is better than uniformly increasing them before tuning.

\noindent \textbf{Effects of early pruning on task accuracy and training memory in LLaMA pruning.}
\cref{fig:llama-init-tradeoff} shows the effect of the initial density on LLaMA models' task performance under the 30\% sparsity pruning setting. We find that densely-trained models only perform better in TruthfulQA with fewer parameters pruned before tuning. The accuracy reaches 48.6 and 47.4 when not pruning before tuning, compared to 46.6 and 44.7 when directly pruning to the target sparsity for both 7B and 13B models. Training the {\lmabbr} densely harms the model performance while costing extra memory for all other tasks. These results demonstrate that pruning during training hurts large LM performance under distillation-free settings, and we hypothesize this is due to the training instability issue when parameters are set to zeros during fine-tuning.

\section{Absolute Efficiency Metrics} \label{sec:raw-efficiency-results}
We report the raw efficiency evaluation results in \cref{tab:raw-efficiency} and \cref{tab:raw-llama-efficiency}, including training and inference time and memory consumption. The training times are measured in seconds, and the inference times are measured in milliseconds. All memory footprints are measured in MB. We report the time-to-accuracy for RoBERTa and T5 model training to measure the training time. For LLaMA model training, we measure the training time per epoch to represent training time consumption. 

\input{tables/raw_roberta_efficiency}

\input{tables/raw_llama_efficiency}

%% file: tables/appendix_hyperparam.tex
\begin{table}[htbp]
\centering
\begin{tabular}{@{}llllll@{}}
\toprule
Hypeparameter  & GLUE-small & GLUE-big & SQuAD & CNN/DM & Alpaca \\ \midrule
Learning rate  & 2e-4       & 2e-4     & 2e-4  & 1e-4   & 1e-4   \\
Batch size     & 32         & 32       & 32    & 16     & 32     \\
Epochs         & 40         & 40       & 40    & 16     & 15     \\
Distill epochs & 20         & 20       & 20    & 6      & -      \\ \bottomrule
\end{tabular}
\caption{Hyperparameters used in {\ourmethod} experiments}
\label{tab:appendix-hyperparam}
\end{table}

%% file: tables/bert_additional_results.tex
\begin{table*}[htbp]
\centering
\resizebox{1.0\textwidth}{!}{
\begin{tabular}{@{}rlrrrrrrrrr@{}}
\toprule
Density               & Method  & MNLI          & QQP           & QNLI          & SST2          & CoLA          & STS-B         & MRPC          & RTE           & GLUE Avg.     \\ \midrule
\multirow{5}{*}{50\%} & MaP & \textbf{83.6} & {\ul 87.8}    & \textbf{91.5} & 91.0          & \textbf{60.1} & \textbf{89.8} & 90.7          & 67.2          & {\ul 82.7}    \\
                      & MvP & 82.3          & 87.3          & {\ul 90.8}    & 90.8          & 57.7          & {\ul 89.4}    & {\ul 91.1}    & 67.2          & 82.1          \\
                      & PST & 81.0            & 85.8          & 89.8          & {\ul 91.3}    & 57.6          & 84.6          & 90.7          & 67.9          & 81.0            \\
                      & LRP & 82.4          & 87.2          & 89.6          & 90.9          & 54.1          & 88.7          & 89.8          & {\ul 69.3}    & 82.2          \\ \cmidrule(l){2-11} 
                      & APT & {\ul 82.8}    & \textbf{90.1} & 90.1          & \textbf{92.7} & {\ul 59.6}    & 88.3          & \textbf{91.8} & \textbf{70.4} & \textbf{83.2} \\ \midrule
\multirow{5}{*}{10\%} & MaP & 78.2          & 83.2          & 84.1          & 85.4          & 27.9          & 82.3          & 80.5          & 50.1          & 71.4          \\
                      & MvP & \textbf{80.1} & 84.4          & \textbf{87.2} & 87.2          & 28.6          & {\ul 84.3}    & 84.1          & 57.6          & 74.2          \\
                      & PST & {\ul 79.6}    & {\ul 86.1}    & {\ul 86.6}    & 89.0            & \textbf{38.0}   & 81.3          & 83.6          & {\ul 63.2}    & {\ul 75.9}    \\
                      & LRP & 79.4          & 86.0            & 85.3          & {\ul 89.1}    & {\ul 35.6}    & 83.3          & {\ul 84.4}    & 62.8          & 75.7          \\ \cmidrule(l){2-11} 
                      & APT & 78.8          & \textbf{89.4} & 85.5          & \textbf{90.0}   & 30.9          & \textbf{86.3} & \textbf{88.2} & \textbf{65.3} & \textbf{76.8} \\ \bottomrule
\end{tabular}
}
\caption{Comparison of {\ourmethod} to existing unstructured pruning baseline with using PEFT in conjunction. The best results are \textbf{bold} while the second-best ones are \ul{underlined}.}
\label{tab:bert-additional-result}
\end{table*}

%% file: tables/roberta_detailed_results.tex
\begin{table*}[htbp]
\centering
\begin{tabular}{ll|rrrrrrrr}
\hline
Sparsity              & Method       & MNLI & QQP  & QNLI & SST2 & CoLA & MRPC & RTE  & GLUE Avg. \\ \hline
\multirow{2}{*}{0\%}  & FT           & 87.6 & 91.9 & 92.8 & 95.2 & 91.2 & 90.2 & 78.7 & 89.7      \\
                      & LoRA         & 87.5 & 90.8 & 93.3 & 95.0 & 63.4 & 89.7 & 72.1 & 84.5      \\ \hline
\multirow{2}{*}{40\%} & LoRA+Distill & 84.2 & 88.3 & 90.1 & 91.9 & 49.9 & 86.8 & 68.6 & 80.0      \\ 
                      & APT          & 86.4 & 90.9 & 92.3 & 94.5 & 56.5 & 92.3 & 74.4 & 83.9      \\ \hline
\end{tabular}
\caption{Detailed results of RoBERTa pruning with {\ourmethod} compared to the LoRA+Distill baseline. We ignore the evaluation results of the STS-B task since it cannot be successfully reproduced with CoFi (the distillation backbone).}
\label{tab:roberta-detailed-results}
\end{table*}

%% file: tables/llama_results.tex
\begin{table}[htbp]
\centering
\begin{tabular}{@{}l|lrrrr@{}}
\toprule
Method                            & ARC           & HellaSwag     & MMLU          & TruthfulQA    & Avg.          \\ \midrule
LLaMA2 7B                              & 53.1          & 77.7          & 43.8          & 39.0          & 53.4          \\ \midrule
LoRA                              & 55.6          & 79.3          & 46.9          & 49.9          & 57.9          \\ \midrule
LoRA+Prune & \textbf{46.8} & 65.2          & 23.9          & 46.2          & 45.5          \\
LLMPruner                         & 39.2          & 67.0          & 24.9          & 40.6          & 42.9          \\
APT                               & 45.4          & \textbf{71.1} & \textbf{36.9} & \textbf{46.6} & \textbf{50.0} \\ \midrule
LLaMA2 13B                       & 59.4          & 82.1          & 55.8          & 37.4          & 58.7          \\ \midrule
LoRA                              & 60.8          & 82.8          & 56.0          & 46.5          & 61.5          \\ \midrule
LoRA+Prune & \textbf{56.4} & \textbf{79.1} & 50.7          & 42.1          & \textbf{57.1} \\
LLMPruner                         & 46.8          & 74.0          & 24.7          & 34.8          & 45.1          \\
APT                               & 49.5          & 75.8          & \textbf{52.5} & \textbf{44.7} & 55.6          \\ \bottomrule
\end{tabular}
\caption{LLaMA2 7B and 13B 30\% sparsity pruning results with GPT4-generated Alpaca dataset, evaluated on the Open LLM leaderboard few-shot tasks.}
\label{tab:llama-results}
\vspace{-10pt}
\end{table}

%% file: tables/ablate_distillation.tex
\begin{table}[htbp]
\centering
\begin{tabular}{lr|rr}
\hline
                        & SST2 & Train. Speed($\Uparrow$)      & Train. Mem.($\Downarrow$)   \\ \hline
\ourmethod                             & 94.5 & 16.9\% & 70.1\% \\
\tableindent w/o $\mathcal{L}_{layer}$ & 93.7 & 17.4\%  & 69.8\% \\
\tableindent w/o self-distillation          & 92.9 & 20.7\%  & 69.2\% \\ \hline
FT teacher         &   94.3  &  7.9\% & 111.8\%  \\
LoRA teacher       &   93.7 &  1.7\%   &  96.1\%\\ \hline
\end{tabular}
\caption{Ablation study of distillation strategies and comparison to non-efficient distillation techniques. The training speed and memory are relative metrics compared to fine-tuning the dense model.}
\label{tab:ablate-distillation}
\vspace{-10pt}
\end{table}

%% file: tables/raw_roberta_efficiency.tex
\begin{table}[htbp]
\centering
\begin{tabular}{@{}llr|rrrr@{}}
\toprule
Model                                           & Method             & Sparsity & 97\% TTA (s) & Train Mem. (MB) & Inf. Time (ms) & Inf. Mem (MB) \\ \midrule
\multirow{6}{*}{$\text{RoBERTa}_{\text{base}}$} & FT                 & 0\%      & 127          & 2,696           & 220.8          & 1,157         \\
                                                & LoRA               & 0\%      & 2,714        & 1,630           & 181.8          & 1,157         \\ \cmidrule(l){2-7} 
                                                & LoRA+Prune         & 60\%     & 6,513        & 1,630           & 84.0           & 869           \\
                                                & Prune+Distill      & 60\%     & 1,899        & 4,544           & 85.2           & 917           \\
                                                & LoRA+Prune+Distill & 60\%     & 8,299        & 3,813           & 87.0           & 952           \\ \cmidrule(l){2-7} 
                                                & APT                & 60\%     & 752          & 1,890           & 91.3           & 904           \\ \midrule
\multirow{4}{*}{$\text{T5}_{\text{base}}$}      & FT                 & 0\%      & 366          & 7,217           & 248.1          & 2,347         \\
                                                & LoRA               & 0\%      & 935          & 4,476           & 254.2          & 2,347         \\ \cmidrule(l){2-7} 
                                                & LoRA+Prune         & 60\%     & 14,417       & 4,476           & 116.8          & 1,724         \\ \cmidrule(l){2-7} 
                                                & APT                & 60\%     & 1,774        & 5,332           & 185.0          & 1,913         \\ \bottomrule
\end{tabular}
\caption{Raw efficiency metrics, including time to accuracy, training peak memory, inference time and memory footprints, when using different methods to fine-tune $\text{RoBERTa}_{\text{base}}$ and $\text{T5}_{\text{base}}$models on SST2.}
\label{tab:raw-efficiency}
\end{table}

%% file: tables/raw_llama_efficiency.tex
\begin{table}[htbp]
\centering
\begin{tabular}{@{}l|rrrr@{}}
\toprule
Method          & Train Time (s) & Train Mem. (MB) & Inf. Time (ms) & Inf. Mem (MB) \\ \midrule
LoRA            & 980            & 32,185          & 2457.5         & 45,311        \\
LoRA+MT         & 980            & 32,185          & 2127.5         & 31,207        \\
LoRA+MT+retrain & 1,773          & 32,185          & 2127.5         & 31,207        \\
LLMPruner       & 852            & 23,425          & 2140.6         & 33,625        \\ \midrule
APT             & 1,039          & 24,408          & 2099.7         & 30,469        \\ \bottomrule
\end{tabular}
\caption{Raw efficiency metrics, including time to accuracy, training peak memory, inference time, and memory footprints, when using different methods to fine-tune LLaMA2 7B models on Alpaca.}
\label{tab:raw-llama-efficiency}
\end{table}

%% file: example_paper.bbl
\begin{thebibliography}{60}
\providecommand{\natexlab}[1]{#1}
\providecommand{\url}[1]{\texttt{#1}}
\expandafter\ifx\csname urlstyle\endcsname\relax
  \providecommand{\doi}[1]{doi: #1}\else
  \providecommand{\doi}{doi: \begingroup \urlstyle{rm}\Url}\fi

\bibitem[Ben~Zaken et~al.(2022)Ben~Zaken, Goldberg, and Ravfogel]{ben-zaken-etal-2022-bitfit}
Ben~Zaken, E., Goldberg, Y., and Ravfogel, S.
\newblock {B}it{F}it: Simple parameter-efficient fine-tuning for transformer-based masked language-models.
\newblock In \emph{Proceedings of the 60th Annual Meeting of the Association for Computational Linguistics (Volume 2: Short Papers)}, pp.\  1--9, Dublin, Ireland, 2022. Association for Computational Linguistics.
\newblock \doi{10.18653/v1/2022.acl-short.1}.

\bibitem[Clark et~al.(2018)Clark, Cowhey, Etzioni, Khot, Sabharwal, Schoenick, and Tafjord]{clark2018think}
Clark, P., Cowhey, I., Etzioni, O., Khot, T., Sabharwal, A., Schoenick, C., and Tafjord, O.
\newblock Think you have solved question answering? try arc, the ai2 reasoning challenge.
\newblock \emph{ArXiv preprint}, abs/1803.05457, 2018.

\bibitem[Coleman et~al.(2019)Coleman, Kang, Narayanan, Nardi, Zhao, Zhang, Bailis, Olukotun, R\'{e}, and Zaharia]{10.1145/3352020.3352024}
Coleman, C., Kang, D., Narayanan, D., Nardi, L., Zhao, T., Zhang, J., Bailis, P., Olukotun, K., R\'{e}, C., and Zaharia, M.
\newblock Analysis of dawnbench, a time-to-accuracy machine learning performance benchmark.
\newblock \emph{SIGOPS Oper. Syst. Rev.}, 53\penalty0 (1):\penalty0 14–25, 2019.
\newblock ISSN 0163-5980.
\newblock \doi{10.1145/3352020.3352024}.

\bibitem[Dettmers et~al.(2022)Dettmers, Lewis, Belkada, and Zettlemoyer]{Dettmers2022LLMint88M}
Dettmers, T., Lewis, M., Belkada, Y., and Zettlemoyer, L.
\newblock Gpt3.int8(): 8-bit matrix multiplication for transformers at scale.
\newblock In Koyejo, S., Mohamed, S., Agarwal, A., Belgrave, D., Cho, K., and Oh, A. (eds.), \emph{Advances in Neural Information Processing Systems}, volume~35, pp.\  30318--30332. Curran Associates, Inc., 2022.

\bibitem[Dettmers et~al.(2023)Dettmers, Pagnoni, Holtzman, and Zettlemoyer]{dettmers2023qlora}
Dettmers, T., Pagnoni, A., Holtzman, A., and Zettlemoyer, L.
\newblock Qlora: Efficient finetuning of quantized llms.
\newblock \emph{ArXiv preprint}, abs/2305.14314, 2023.

\bibitem[Devlin et~al.(2019)Devlin, Chang, Lee, and Toutanova]{devlin-etal-2019-bert}
Devlin, J., Chang, M.-W., Lee, K., and Toutanova, K.
\newblock {BERT}: Pre-training of deep bidirectional transformers for language understanding.
\newblock In \emph{Proceedings of the 2019 Conference of the North {A}merican Chapter of the Association for Computational Linguistics: Human Language Technologies, Volume 1 (Long and Short Papers)}, pp.\  4171--4186, Minneapolis, Minnesota, 2019. Association for Computational Linguistics.
\newblock \doi{10.18653/v1/N19-1423}.

\bibitem[Ding et~al.(2023)Ding, Qin, Yang, Wei, Yang, Su, Hu, Chen, Chan, Chen, et~al.]{ding2023parameter}
Ding, N., Qin, Y., Yang, G., Wei, F., Yang, Z., Su, Y., Hu, S., Chen, Y., Chan, C.-M., Chen, W., et~al.
\newblock Parameter-efficient fine-tuning of large-scale pre-trained language models.
\newblock \emph{Nature Machine Intelligence}, 5\penalty0 (3):\penalty0 220--235, 2023.

\bibitem[Frankle \& Carbin(2019)Frankle and Carbin]{frankle2018lottery}
Frankle, J. and Carbin, M.
\newblock The lottery ticket hypothesis: Finding sparse, trainable neural networks.
\newblock In \emph{7th International Conference on Learning Representations, {ICLR} 2019, New Orleans, LA, USA, May 6-9, 2019}. OpenReview.net, 2019.

\bibitem[Frankle et~al.(2021)Frankle, Dziugaite, Roy, and Carbin]{frankle2021pruning}
Frankle, J., Dziugaite, G.~K., Roy, D., and Carbin, M.
\newblock Pruning neural networks at initialization: Why are we missing the mark?
\newblock In \emph{9th International Conference on Learning Representations, {ICLR} 2021, Virtual Event, Austria, May 3-7, 2021}. OpenReview.net, 2021.

\bibitem[Frantar \& Alistarh(2023)Frantar and Alistarh]{Frantar2023SparseGPTML}
Frantar, E. and Alistarh, D.
\newblock {S}parse{GPT}: Massive language models can be accurately pruned in one-shot.
\newblock In Krause, A., Brunskill, E., Cho, K., Engelhardt, B., Sabato, S., and Scarlett, J. (eds.), \emph{Proceedings of the 40th International Conference on Machine Learning}, volume 202 of \emph{Proceedings of Machine Learning Research}, pp.\  10323--10337. PMLR, 2023.

\bibitem[Frantar et~al.(2023)Frantar, Ashkboos, Hoefler, and Alistarh]{Frantar2022GPTQAP}
Frantar, E., Ashkboos, S., Hoefler, T., and Alistarh, D.
\newblock {OPTQ}: Accurate quantization for generative pre-trained transformers.
\newblock In \emph{The Eleventh International Conference on Learning Representations}, 2023.

\bibitem[Gao et~al.(2023)Gao, Tow, Abbasi, Biderman, Black, DiPofi, Foster, Golding, Hsu, Le~Noac'h, Li, McDonell, Muennighoff, Ociepa, Phang, Reynolds, Schoelkopf, Skowron, Sutawika, Tang, Thite, Wang, Wang, and Zou]{eval-harness}
Gao, L., Tow, J., Abbasi, B., Biderman, S., Black, S., DiPofi, A., Foster, C., Golding, L., Hsu, J., Le~Noac'h, A., Li, H., McDonell, K., Muennighoff, N., Ociepa, C., Phang, J., Reynolds, L., Schoelkopf, H., Skowron, A., Sutawika, L., Tang, E., Thite, A., Wang, B., Wang, K., and Zou, A.
\newblock A framework for few-shot language model evaluation, 2023.

\bibitem[Guo et~al.(2021)Guo, Rush, and Kim]{guo-etal-2021-parameter}
Guo, D., Rush, A., and Kim, Y.
\newblock Parameter-efficient transfer learning with diff pruning.
\newblock In \emph{Proceedings of the 59th Annual Meeting of the Association for Computational Linguistics and the 11th International Joint Conference on Natural Language Processing (Volume 1: Long Papers)}, pp.\  4884--4896, Online, 2021. Association for Computational Linguistics.
\newblock \doi{10.18653/v1/2021.acl-long.378}.

\bibitem[Haidar et~al.(2022)Haidar, Anchuri, Rezagholizadeh, Ghaddar, Langlais, and Poupart]{haidar2022rail}
Haidar, M.~A., Anchuri, N., Rezagholizadeh, M., Ghaddar, A., Langlais, P., and Poupart, P.
\newblock {RAIL}-{KD}: {RA}ndom intermediate layer mapping for knowledge distillation.
\newblock In \emph{Findings of the Association for Computational Linguistics: NAACL 2022}, pp.\  1389--1400, Seattle, United States, 2022. Association for Computational Linguistics.
\newblock \doi{10.18653/v1/2022.findings-naacl.103}.

\bibitem[Han et~al.(2015)Han, Pool, Tran, and Dally]{han2015learning}
Han, S., Pool, J., Tran, J., and Dally, W.~J.
\newblock Learning both weights and connections for efficient neural network.
\newblock In Cortes, C., Lawrence, N.~D., Lee, D.~D., Sugiyama, M., and Garnett, R. (eds.), \emph{Advances in Neural Information Processing Systems 28: Annual Conference on Neural Information Processing Systems 2015, December 7-12, 2015, Montreal, Quebec, Canada}, pp.\  1135--1143, 2015.

\bibitem[Han et~al.(2016)Han, Mao, and Dally]{Han2015DeepCC}
Han, S., Mao, H., and Dally, W.~J.
\newblock Deep compression: Compressing deep neural network with pruning, trained quantization and huffman coding.
\newblock In Bengio, Y. and LeCun, Y. (eds.), \emph{4th International Conference on Learning Representations, {ICLR} 2016, San Juan, Puerto Rico, May 2-4, 2016, Conference Track Proceedings}, 2016.

\bibitem[He et~al.(2022{\natexlab{a}})He, Zhou, Ma, Berg{-}Kirkpatrick, and Neubig]{he_towards_2022}
He, J., Zhou, C., Ma, X., Berg{-}Kirkpatrick, T., and Neubig, G.
\newblock Towards a unified view of parameter-efficient transfer learning.
\newblock In \emph{The Tenth International Conference on Learning Representations, {ICLR} 2022, Virtual Event, April 25-29, 2022}. OpenReview.net, 2022{\natexlab{a}}.

\bibitem[He et~al.(2022{\natexlab{b}})He, Ding, Dong, Zhang, and Tao]{he-etal-2022-sparseadapter}
He, S., Ding, L., Dong, D., Zhang, J., and Tao, D.
\newblock {S}parse{A}dapter: An easy approach for improving the parameter-efficiency of adapters.
\newblock In \emph{Findings of the Association for Computational Linguistics: EMNLP 2022}, pp.\  2184--2190, Abu Dhabi, United Arab Emirates, 2022{\natexlab{b}}. Association for Computational Linguistics.

\bibitem[Hedegaard et~al.(2022)Hedegaard, Alok, Jose, and Iosifidis]{hedegaard_structured_2022}
Hedegaard, L., Alok, A., Jose, J., and Iosifidis, A.
\newblock Structured {Pruning} {Adapters}, 2022.

\bibitem[Hendrycks et~al.(2021)Hendrycks, Burns, Basart, Zou, Mazeika, Song, and Steinhardt]{hendrycks2021measuring}
Hendrycks, D., Burns, C., Basart, S., Zou, A., Mazeika, M., Song, D., and Steinhardt, J.
\newblock Measuring massive multitask language understanding.
\newblock In \emph{9th International Conference on Learning Representations, {ICLR} 2021, Virtual Event, Austria, May 3-7, 2021}. OpenReview.net, 2021.

\bibitem[Hinton et~al.(2015)Hinton, Vinyals, and Dean]{Hinton2015DistillingTK}
Hinton, G.~E., Vinyals, O., and Dean, J.
\newblock Distilling the knowledge in a neural network.
\newblock \emph{ArXiv preprint}, abs/1503.02531, 2015.

\bibitem[Houlsby et~al.(2019)Houlsby, Giurgiu, Jastrzebski, Morrone, de~Laroussilhe, Gesmundo, Attariyan, and Gelly]{houlsby2019parameter}
Houlsby, N., Giurgiu, A., Jastrzebski, S., Morrone, B., de~Laroussilhe, Q., Gesmundo, A., Attariyan, M., and Gelly, S.
\newblock Parameter-efficient transfer learning for {NLP}.
\newblock In Chaudhuri, K. and Salakhutdinov, R. (eds.), \emph{Proceedings of the 36th International Conference on Machine Learning, {ICML} 2019, 9-15 June 2019, Long Beach, California, {USA}}, volume~97 of \emph{Proceedings of Machine Learning Research}, pp.\  2790--2799. {PMLR}, 2019.

\bibitem[Hu et~al.(2022)Hu, Shen, Wallis, Allen{-}Zhu, Li, Wang, Wang, and Chen]{hu_lora_2021}
Hu, E.~J., Shen, Y., Wallis, P., Allen{-}Zhu, Z., Li, Y., Wang, S., Wang, L., and Chen, W.
\newblock Lora: Low-rank adaptation of large language models.
\newblock In \emph{The Tenth International Conference on Learning Representations, {ICLR} 2022, Virtual Event, April 25-29, 2022}. OpenReview.net, 2022.

\bibitem[Kaplan et~al.(2020)Kaplan, McCandlish, Henighan, Brown, Chess, Child, Gray, Radford, Wu, and Amodei]{kaplan2020scaling}
Kaplan, J., McCandlish, S., Henighan, T., Brown, T.~B., Chess, B., Child, R., Gray, S., Radford, A., Wu, J., and Amodei, D.
\newblock Scaling laws for neural language models.
\newblock \emph{ArXiv preprint}, abs/2001.08361, 2020.

\bibitem[Kwon et~al.(2022)Kwon, Kim, Mahoney, Hassoun, Keutzer, and Gholami]{kwon_fast_2022}
Kwon, W., Kim, S., Mahoney, M.~W., Hassoun, J., Keutzer, K., and Gholami, A.
\newblock A fast post-training pruning framework for transformers.
\newblock In Koyejo, S., Mohamed, S., Agarwal, A., Belgrave, D., Cho, K., and Oh, A. (eds.), \emph{Advances in Neural Information Processing Systems}, volume~35, pp.\  24101--24116. Curran Associates, Inc., 2022.

\bibitem[Lagunas et~al.(2021)Lagunas, Charlaix, Sanh, and Rush]{lagunas_block_2021}
Lagunas, F., Charlaix, E., Sanh, V., and Rush, A.
\newblock Block pruning for faster transformers.
\newblock In \emph{Proceedings of the 2021 Conference on Empirical Methods in Natural Language Processing}, pp.\  10619--10629, Online and Punta Cana, Dominican Republic, 2021. Association for Computational Linguistics.
\newblock \doi{10.18653/v1/2021.emnlp-main.829}.

\bibitem[LeCun et~al.(1989)LeCun, Denker, and Solla]{LeCun1989OptimalBD}
LeCun, Y., Denker, J.~S., and Solla, S.~A.
\newblock Optimal brain damage.
\newblock In \emph{NIPS}, 1989.

\bibitem[Lester et~al.(2021)Lester, Al-Rfou, and Constant]{lester-etal-2021-power}
Lester, B., Al-Rfou, R., and Constant, N.
\newblock The power of scale for parameter-efficient prompt tuning.
\newblock In \emph{Proceedings of the 2021 Conference on Empirical Methods in Natural Language Processing}, pp.\  3045--3059, Online and Punta Cana, Dominican Republic, 2021. Association for Computational Linguistics.
\newblock \doi{10.18653/v1/2021.emnlp-main.243}.

\bibitem[Li \& Liang(2021)Li and Liang]{li2021prefix}
Li, X.~L. and Liang, P.
\newblock Prefix-tuning: Optimizing continuous prompts for generation.
\newblock In \emph{Proceedings of the 59th Annual Meeting of the Association for Computational Linguistics and the 11th International Joint Conference on Natural Language Processing (Volume 1: Long Papers)}, pp.\  4582--4597, Online, 2021. Association for Computational Linguistics.
\newblock \doi{10.18653/v1/2021.acl-long.353}.

\bibitem[Li et~al.(2022)Li, Luo, Tan, Wang, Huang, Li, and Bai]{ijcai2022p586}
Li, Y., Luo, F., Tan, C., Wang, M., Huang, S., Li, S., and Bai, J.
\newblock Parameter-efficient sparsity for large language models fine-tuning.
\newblock In Raedt, L.~D. (ed.), \emph{Proceedings of the Thirty-First International Joint Conference on Artificial Intelligence, {IJCAI-22}}, pp.\  4223--4229. International Joint Conferences on Artificial Intelligence Organization, 2022.
\newblock \doi{10.24963/ijcai.2022/586}.
\newblock Main Track.

\bibitem[Lialin et~al.(2023)Lialin, Deshpande, and Rumshisky]{lialin2023scaling}
Lialin, V., Deshpande, V., and Rumshisky, A.
\newblock Scaling down to scale up: A guide to parameter-efficient fine-tuning.
\newblock \emph{ArXiv preprint}, abs/2303.15647, 2023.

\bibitem[Lin et~al.(2023)Lin, Tang, Tang, Yang, Dang, and Han]{lin2023awq}
Lin, J., Tang, J., Tang, H., Yang, S., Dang, X., and Han, S.
\newblock Awq: Activation-aware weight quantization for llm compression and acceleration.
\newblock \emph{ArXiv preprint}, abs/2306.00978, 2023.

\bibitem[Lin et~al.(2022)Lin, Hilton, and Evans]{lin-etal-2022-truthfulqa}
Lin, S., Hilton, J., and Evans, O.
\newblock {T}ruthful{QA}: Measuring how models mimic human falsehoods.
\newblock In \emph{Proceedings of the 60th Annual Meeting of the Association for Computational Linguistics (Volume 1: Long Papers)}, pp.\  3214--3252, Dublin, Ireland, 2022. Association for Computational Linguistics.
\newblock \doi{10.18653/v1/2022.acl-long.229}.

\bibitem[Liu et~al.(2019)Liu, Ott, Goyal, Du, Joshi, Chen, Levy, Lewis, Zettlemoyer, and Stoyanov]{liu2019roberta}
Liu, Y., Ott, M., Goyal, N., Du, J., Joshi, M., Chen, D., Levy, O., Lewis, M., Zettlemoyer, L., and Stoyanov, V.
\newblock Roberta: A robustly optimized bert pretraining approach.
\newblock \emph{ArXiv preprint}, abs/1907.11692, 2019.

\bibitem[Ma et~al.(2023)Ma, Fang, and Wang]{ma2023llm}
Ma, X., Fang, G., and Wang, X.
\newblock Llm-pruner: On the structural pruning of large language models.
\newblock \emph{ArXiv preprint}, abs/2305.11627, 2023.

\bibitem[Mahabadi et~al.(2021)Mahabadi, Henderson, and Ruder]{karimi2021compacter}
Mahabadi, R.~K., Henderson, J., and Ruder, S.
\newblock Compacter: Efficient low-rank hypercomplex adapter layers.
\newblock In Ranzato, M., Beygelzimer, A., Dauphin, Y.~N., Liang, P., and Vaughan, J.~W. (eds.), \emph{Advances in Neural Information Processing Systems 34: Annual Conference on Neural Information Processing Systems 2021, NeurIPS 2021, December 6-14, 2021, virtual}, pp.\  1022--1035, 2021.

\bibitem[Mishra et~al.(2022)Mishra, Khashabi, Baral, and Hajishirzi]{mishra-etal-2022-cross}
Mishra, S., Khashabi, D., Baral, C., and Hajishirzi, H.
\newblock Cross-task generalization via natural language crowdsourcing instructions.
\newblock In \emph{Proceedings of the 60th Annual Meeting of the Association for Computational Linguistics (Volume 1: Long Papers)}, pp.\  3470--3487, Dublin, Ireland, 2022. Association for Computational Linguistics.
\newblock \doi{10.18653/v1/2022.acl-long.244}.

\bibitem[Nallapati et~al.(2016)Nallapati, Zhou, dos Santos, Gulcehre, and Xiang]{Nallapati2016AbstractiveTS}
Nallapati, R., Zhou, B., dos Santos, C., Gulcehre, C., and Xiang, B.
\newblock Abstractive text summarization using sequence-to-sequence {RNN}s and beyond.
\newblock In \emph{Proceedings of the 20th {SIGNLL} Conference on Computational Natural Language Learning}, pp.\  280--290, Berlin, Germany, 2016. Association for Computational Linguistics.
\newblock \doi{10.18653/v1/K16-1028}.

\bibitem[Panigrahi et~al.(2023)Panigrahi, Saunshi, Zhao, and Arora]{panigrahi2023task}
Panigrahi, A., Saunshi, N., Zhao, H., and Arora, S.
\newblock Task-specific skill localization in fine-tuned language models.
\newblock \emph{ArXiv preprint}, abs/2302.06600, 2023.

\bibitem[Pfeiffer et~al.(2021)Pfeiffer, Kamath, R{\"u}ckl{\'e}, Cho, and Gurevych]{Pfeiffer2020AdapterFusionNT}
Pfeiffer, J., Kamath, A., R{\"u}ckl{\'e}, A., Cho, K., and Gurevych, I.
\newblock {A}dapter{F}usion: Non-destructive task composition for transfer learning.
\newblock In \emph{Proceedings of the 16th Conference of the European Chapter of the Association for Computational Linguistics: Main Volume}, pp.\  487--503, Online, 2021. Association for Computational Linguistics.
\newblock \doi{10.18653/v1/2021.eacl-main.39}.

\bibitem[Raffel et~al.(2020)Raffel, Shazeer, Roberts, Lee, Narang, Matena, Zhou, Li, and Liu]{raffel2020exploring}
Raffel, C., Shazeer, N., Roberts, A., Lee, K., Narang, S., Matena, M., Zhou, Y., Li, W., and Liu, P.~J.
\newblock Exploring the limits of transfer learning with a unified text-to-text transformer.
\newblock \emph{J. Mach. Learn. Res.}, 21:\penalty0 140:1--140:67, 2020.

\bibitem[Rajpurkar et~al.(2018)Rajpurkar, Jia, and Liang]{rajpurkar-etal-2018-know}
Rajpurkar, P., Jia, R., and Liang, P.
\newblock Know what you don{'}t know: Unanswerable questions for {SQ}u{AD}.
\newblock In \emph{Proceedings of the 56th Annual Meeting of the Association for Computational Linguistics (Volume 2: Short Papers)}, pp.\  784--789, Melbourne, Australia, 2018. Association for Computational Linguistics.
\newblock \doi{10.18653/v1/P18-2124}.

\bibitem[Sanh et~al.(2020)Sanh, Wolf, and Rush]{sanh_movement_2020}
Sanh, V., Wolf, T., and Rush, A.~M.
\newblock Movement pruning: Adaptive sparsity by fine-tuning.
\newblock In Larochelle, H., Ranzato, M., Hadsell, R., Balcan, M., and Lin, H. (eds.), \emph{Advances in Neural Information Processing Systems 33: Annual Conference on Neural Information Processing Systems 2020, NeurIPS 2020, December 6-12, 2020, virtual}, 2020.

\bibitem[Shen et~al.(2022{\natexlab{a}})Shen, Molchanov, Yin, and Alvarez]{Shen_2022_CVPR}
Shen, M., Molchanov, P., Yin, H., and Alvarez, J.~M.
\newblock When to prune? a policy towards early structural pruning.
\newblock In \emph{Proceedings of the IEEE/CVF Conference on Computer Vision and Pattern Recognition (CVPR)}, pp.\  12247--12256, 2022{\natexlab{a}}.

\bibitem[Shen et~al.(2022{\natexlab{b}})Shen, Yin, Molchanov, Mao, Liu, and Alvarez]{NEURIPS2022_5434be94}
Shen, M., Yin, H., Molchanov, P., Mao, L., Liu, J., and Alvarez, J.~M.
\newblock Structural pruning via latency-saliency knapsack.
\newblock In Koyejo, S., Mohamed, S., Agarwal, A., Belgrave, D., Cho, K., and Oh, A. (eds.), \emph{Advances in Neural Information Processing Systems}, volume~35, pp.\  12894--12908. Curran Associates, Inc., 2022{\natexlab{b}}.

\bibitem[Sun et~al.(2023)Sun, Liu, Bair, and Kolter]{sun2023simple}
Sun, M., Liu, Z., Bair, A., and Kolter, J.~Z.
\newblock A simple and effective pruning approach for large language models.
\newblock \emph{ArXiv preprint}, abs/2306.11695, 2023.

\bibitem[Sung et~al.(2021)Sung, Nair, and Raffel]{sung2021training}
Sung, Y., Nair, V., and Raffel, C.
\newblock Training neural networks with fixed sparse masks.
\newblock In Ranzato, M., Beygelzimer, A., Dauphin, Y.~N., Liang, P., and Vaughan, J.~W. (eds.), \emph{Advances in Neural Information Processing Systems 34: Annual Conference on Neural Information Processing Systems 2021, NeurIPS 2021, December 6-14, 2021, virtual}, pp.\  24193--24205, 2021.

\bibitem[Taori et~al.(2023)Taori, Gulrajani, Zhang, Dubois, Li, Guestrin, Liang, and Hashimoto]{alpaca}
Taori, R., Gulrajani, I., Zhang, T., Dubois, Y., Li, X., Guestrin, C., Liang, P., and Hashimoto, T.~B.
\newblock Stanford alpaca: An instruction-following llama model.
\newblock \url{https://github.com/tatsu-lab/stanford_alpaca}, 2023.

\bibitem[Touvron et~al.(2023)Touvron, Lavril, Izacard, Martinet, Lachaux, Lacroix, Rozi{\`e}re, Goyal, Hambro, Azhar, et~al.]{touvron2023llama}
Touvron, H., Lavril, T., Izacard, G., Martinet, X., Lachaux, M.-A., Lacroix, T., Rozi{\`e}re, B., Goyal, N., Hambro, E., Azhar, F., et~al.
\newblock Llama: Open and efficient foundation language models.
\newblock \emph{ArXiv preprint}, abs/2302.13971, 2023.

\bibitem[Wang et~al.(2019)Wang, Singh, Michael, Hill, Levy, and Bowman]{wang2018glue}
Wang, A., Singh, A., Michael, J., Hill, F., Levy, O., and Bowman, S.~R.
\newblock {GLUE:} {A} multi-task benchmark and analysis platform for natural language understanding.
\newblock In \emph{7th International Conference on Learning Representations, {ICLR} 2019, New Orleans, LA, USA, May 6-9, 2019}. OpenReview.net, 2019.

\bibitem[Wang et~al.(2022{\natexlab{a}})Wang, Wen, Zhang, Hou, Liu, and Li]{wang-etal-2022-finding-skill}
Wang, X., Wen, K., Zhang, Z., Hou, L., Liu, Z., and Li, J.
\newblock Finding skill neurons in pre-trained transformer-based language models.
\newblock In \emph{Proceedings of the 2022 Conference on Empirical Methods in Natural Language Processing}, pp.\  11132--11152, Abu Dhabi, United Arab Emirates, 2022{\natexlab{a}}. Association for Computational Linguistics.

\bibitem[Wang et~al.(2022{\natexlab{b}})Wang, Mishra, Alipoormolabashi, Kordi, Mirzaei, Naik, Ashok, Dhanasekaran, Arunkumar, Stap, Pathak, Karamanolakis, Lai, Purohit, Mondal, Anderson, Kuznia, Doshi, Pal, Patel, Moradshahi, Parmar, Purohit, Varshney, Kaza, Verma, Puri, Karia, Doshi, Sampat, Mishra, Reddy~A, Patro, Dixit, and Shen]{wang-etal-2022-super}
Wang, Y., Mishra, S., Alipoormolabashi, P., Kordi, Y., Mirzaei, A., Naik, A., Ashok, A., Dhanasekaran, A.~S., Arunkumar, A., Stap, D., Pathak, E., Karamanolakis, G., Lai, H., Purohit, I., Mondal, I., Anderson, J., Kuznia, K., Doshi, K., Pal, K.~K., Patel, M., Moradshahi, M., Parmar, M., Purohit, M., Varshney, N., Kaza, P.~R., Verma, P., Puri, R.~S., Karia, R., Doshi, S., Sampat, S.~K., Mishra, S., Reddy~A, S., Patro, S., Dixit, T., and Shen, X.
\newblock Super-{N}atural{I}nstructions: Generalization via declarative instructions on 1600+ {NLP} tasks.
\newblock In \emph{Proceedings of the 2022 Conference on Empirical Methods in Natural Language Processing}, pp.\  5085--5109, Abu Dhabi, United Arab Emirates, 2022{\natexlab{b}}. Association for Computational Linguistics.

\bibitem[Xia et~al.(2022)Xia, Zhong, and Chen]{xia_structured_2022}
Xia, M., Zhong, Z., and Chen, D.
\newblock Structured pruning learns compact and accurate models.
\newblock In \emph{Proceedings of the 60th Annual Meeting of the Association for Computational Linguistics (Volume 1: Long Papers)}, pp.\  1513--1528, Dublin, Ireland, 2022. Association for Computational Linguistics.
\newblock \doi{10.18653/v1/2022.acl-long.107}.

\bibitem[Xu et~al.(2021)Xu, Yen, Zhao, and Xiao]{xu2021rethinking}
Xu, D., Yen, I. E.-H., Zhao, J., and Xiao, Z.
\newblock Rethinking network pruning {--} under the pre-train and fine-tune paradigm.
\newblock In \emph{Proceedings of the 2021 Conference of the North American Chapter of the Association for Computational Linguistics: Human Language Technologies}, pp.\  2376--2382, Online, 2021. Association for Computational Linguistics.
\newblock \doi{10.18653/v1/2021.naacl-main.188}.

\bibitem[Xu et~al.(2023)Xu, Xie, Gu, Chen, Chang, Zhang, Chen, Zhang, and Tian]{xu2023qa}
Xu, Y., Xie, L., Gu, X., Chen, X., Chang, H., Zhang, H., Chen, Z., Zhang, X., and Tian, Q.
\newblock Qa-lora: Quantization-aware low-rank adaptation of large language models.
\newblock \emph{ArXiv preprint}, abs/2309.14717, 2023.

\bibitem[Zellers et~al.(2019)Zellers, Holtzman, Bisk, Farhadi, and Choi]{zellers-etal-2019-hellaswag}
Zellers, R., Holtzman, A., Bisk, Y., Farhadi, A., and Choi, Y.
\newblock {H}ella{S}wag: Can a machine really finish your sentence?
\newblock In \emph{Proceedings of the 57th Annual Meeting of the Association for Computational Linguistics}, pp.\  4791--4800, Florence, Italy, 2019. Association for Computational Linguistics.
\newblock \doi{10.18653/v1/P19-1472}.

\bibitem[Zhang et~al.(2023{\natexlab{a}})Zhang, Shen, Yang, Ou, Yu, Zhuang, et~al.]{zhang2023pruning}
Zhang, M., Shen, C., Yang, Z., Ou, L., Yu, X., Zhuang, B., et~al.
\newblock Pruning meets low-rank parameter-efficient fine-tuning.
\newblock \emph{ArXiv preprint}, abs/2305.18403, 2023{\natexlab{a}}.

\bibitem[Zhang et~al.(2023{\natexlab{b}})Zhang, Chen, Bukharin, He, Cheng, Chen, and Zhao]{zhang2023adaptive}
Zhang, Q., Chen, M., Bukharin, A., He, P., Cheng, Y., Chen, W., and Zhao, T.
\newblock Adaptive budget allocation for parameter-efficient fine-tuning.
\newblock In \emph{The Eleventh International Conference on Learning Representations}, 2023{\natexlab{b}}.

\bibitem[Zhang et~al.(2023{\natexlab{c}})Zhang, Zeng, Lin, Xiao, Wang, Han, Liu, Xie, Sun, and Zhou]{zhang2023emergent}
Zhang, Z., Zeng, Z., Lin, Y., Xiao, C., Wang, X., Han, X., Liu, Z., Xie, R., Sun, M., and Zhou, J.
\newblock Emergent modularity in pre-trained transformers.
\newblock \emph{ArXiv preprint}, abs/2305.18390, 2023{\natexlab{c}}.

\bibitem[Zhao et~al.(2023)Zhao, Huang, Han, Liu, Zhang, and Sun]{zhao2023cpet}
Zhao, W., Huang, Y., Han, X., Liu, Z., Zhang, Z., and Sun, M.
\newblock Cpet: Effective parameter-efficient tuning for compressed large language models.
\newblock \emph{ArXiv preprint}, abs/2307.07705, 2023.

\end{thebibliography}
